\documentclass[runningheads]{llncs}
\usepackage{graphicx}

\usepackage[pagebackref=true,breaklinks=true,colorlinks,bookmarks=false]{hyperref}
\usepackage{tikz}
\usepackage{comment}
\usepackage{amsmath,amssymb} 
\usepackage{eucal}
\usepackage{amssymb}
\usepackage{epsfig}
\usepackage{enumerate}
\usepackage{subcaption}
\usepackage{graphicx}
\usepackage{color}
\usepackage{colortbl}

\usepackage{threeparttable}
\usepackage{multirow,booktabs,makecell}

\usepackage{xspace}
\usepackage{xcolor}

\def\onedot{\ifx\lettoken.\else.\null\fi\xspace}
 
\newcommand{\etal}{et al.\@\xspace}

\newcommand{\backwardwarp}{\overleftarrow{\mathcal{W}}}

\begin{document}
\definecolor{MyRed}{rgb}{0.8,0.2,0}
\definecolor{MyBlue}{rgb}{0,0,1.0}
\def\red#1{\textcolor{MyRed}{#1}}
\def\blue#1{\textcolor{MyBlue}{#1}}
\def\first#1{\red{\textbf{#1}}}
\def\second#1{\blue{\underline{#1}}}
    \pagestyle{headings}
\mainmatter
\def\ECCVSubNumber{95}

\title{Real-Time Intermediate Flow Estimation for Video Frame Interpolation}
\titlerunning{Real-Time Intermediate Flow Estimation for Video Frame Interpolation}

\newcommand*\samethanks[1][\value{footnote}]{\footnotemark[#1]}
\authorrunning{Z. Huang \etal}
\author{Zhewei Huang\inst{1}
		~~
		Tianyuan Zhang\inst{1}
		~~
		Wen Heng\inst{1}
		~~
		Boxin Shi\inst{2,3, 4,}\thanks{Corresponding authors.}
		~~
		Shuchang Zhou\inst{1,}\samethanks
		}
\institute{
		Megvii Research\and NERCVT, School of Computer Science, Peking University\and Institute for Artificial Intelligence, Peking University\and Beijing Academy of Artificial Intelligence\\	\email{\tt\small \{huangzhewei, zhangtianyuan, hengwen, zsc\}@megvii.com, shiboxin@pku.edu.cn}
		\\
		\url{https://github.com/megvii-research/ECCV2022-RIFE}
		}
\maketitle
\begin{abstract}
Real-time video frame interpolation (VFI) is very useful in video processing, media players, and display devices. We propose RIFE, a Real-time Intermediate Flow Estimation algorithm for VFI. To realize a high-quality flow-based VFI method, RIFE uses a neural network named IFNet that can estimate the intermediate flows end-to-end with much faster speed. A privileged distillation scheme is designed for stable IFNet training and improve the overall performance. RIFE does not rely on pre-trained optical flow models and can support arbitrary-timestep frame interpolation with the temporal encoding input. Experiments demonstrate that RIFE achieves state-of-the-art performance on several public benchmarks. Compared with the popular SuperSlomo and DAIN methods, RIFE is 4--27 times faster and produces better results. Furthermore, RIFE can be extended to wider applications thanks to temporal encoding. 
\end{abstract}

\section{Introduction}
Video Frame Interpolation (VFI) aims to synthesize intermediate frames between two consecutive video frames. VFI supports various applications like slow-motion generation, video compression~\cite{wu2018video}, and video frame predition~\cite{wu2022optimizing}. Moreover, real-time VFI methods running on high-resolution videos have many potential applications, such as reducing bandwidth requirements for live video streaming, providing video editing services for users with limited computing resources, and video frame rate adaption on display devices.

VFI is challenging due to the complex, non-linear motions and illumination changes in real-world videos. Recently, flow-based VFI algorithms have offered a framework to address these challenges and achieved impressive results~\cite{liu2017video,jiang2018super,niklaus2018context,xue2019video,bao2019depth,xu2019quadratic,liu2020enhanced}. Common approaches for these methods involve two steps: 1) warping the input frames according to approximated optical flows and 2) fusing the warped frames using Convolutional Neural Networks (CNNs).

Optical flow models can not be directly used in VFI. Given the input frames $I_0, I_1$, flow-based methods~\cite{liu2017video,jiang2018super,bao2019depth} need to approximate the intermediate flows $F_{t\rightarrow 0}, F_{t\rightarrow 1}$ from the perspective of the frame $I_t$ that we are expected to synthesize. There is a ``chicken-and-egg" problem between intermediate flows and frames because $I_t$ is not available beforehand, and its estimation is a difficult problem~\cite{jiang2018super,park2020bmbc}. Many practices~\cite{jiang2018super,bao2019depth,xu2019quadratic,liu2020enhanced} first compute bi-directional flows from optical flow models, then reverse and refine them to generate intermediate flows. However, such flows may have flaws in motion boundaries, as the object position changes from frame to frame~(``object shift" problem). Appearance Flow~\cite{zhou2016view}, A pioneering work in view synthesis, proposes to estimate flow starting from the target view using CNNs. DVF~\cite{liu2017video} extend it to the voxel flow of dynamic scenes to jointly model the intermediate flow and blend mask to estimate them end-to-end. AdaCoF~\cite{lee2020adacof} further extends intermediate flows to adaptive collaborative flows. BMBC~\cite{park2020bmbc} designs a bilateral cost volume operator for obtaining more accurate intermediate flows~(bilateral motion). 
In this paper, we aim to build a lightweight pipeline that achieves state-of-the-art~(SOTA) performance while maintaining the conciseness of direct intermediate flow estimation. Our pipeline has these main design concepts: 
\begin{enumerate}[1)] 
	\item Not requiring additional components, like image depth model~\cite{bao2019depth}, flow refinement model~\cite{jiang2018super} and flow reversal layer~\cite{xu2019quadratic}, which are introduced to compensate for the defects of intermediate flow estimation. We also want to eliminate reliance on pre-trained SOTA optical flow models that are not tailored for VFI tasks. 
	\item End-to-end learnable motion estimation: we demonstrate experimentally that instead of introducing some inaccurate motion modeling, it is better to make the CNN learn the intermediate flow end-to-end. This methodology has been proposed~\cite{liu2017video}. However, the follow-up works do not fully inherit this idea.
	\item Providing direct supervision for the approximated intermediate flows: most VFI models are trained with only the final reconstruction loss. Intuitively, propagating gradients of pixel-wise loss across warping operator is not efficient for flow estimation~\cite{danier2021spatio,meister2017unflow,luo2020upflow}. Lacking supervision explicitly designed for flow estimation degrades the performance of VFI models. 
\end{enumerate}

We propose IFNet, which directly estimates intermediate flow from adjacent frames and a temporal encoding input. IFNet adopts a coarse-to-fine strategy~\cite{ilg2017flownet} with progressively increasing resolution: it iteratively updates the intermediate flows and soft fusion mask via successive IFBlocks. Intuitively, according to the iteratively updated flow fields, we could move corresponding pixels from two input frames to the same location in a latent intermediate frame and use a fusion mask to combine pixels from two input frames. To make our model efficient, unlike most previous optical flow models~\cite{dosovitskiy2015flownet,ilg2017flownet,sun2018pwc,hui2018liteflownet,teed2020raft}, IFBlocks do not contain expensive operators like cost volume and only use $3\times3$ convolution and deconvolution as building blocks, which are suitable for resource-constrained devices. Furthermore, plain Conv is highly supported by NPU embedded in display devices and provides convenience for customized requirements. Thanks related researchers for the exploration of efficient models~\cite{ranjan2017optical,ma2018shufflenet,ding2021repvgg}.


Employing intermediate supervision is very important. When training the IFNet end-to-end using the final reconstruction loss, our method produces worse results than SOTA methods because of the inaccurate optical flow estimation. The situation dramatically changes after we design a privileged distillation scheme that employs a teacher model with access to the intermediate frames to guide the student to learn.

\begin{figure}[tb]
\vspace{-2em}
	\centering
	\includegraphics[width=6cm]{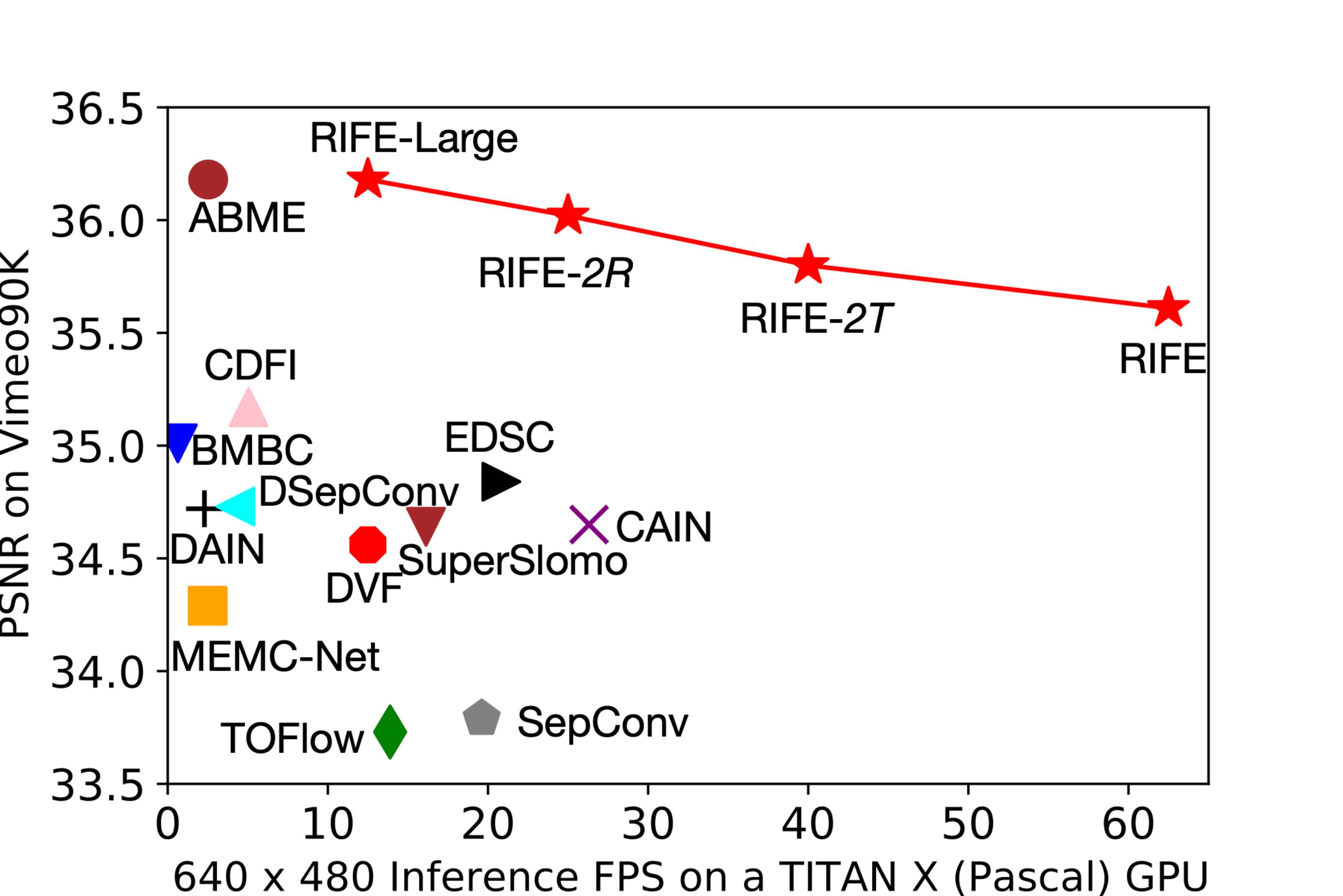}
	\includegraphics[width=6cm]{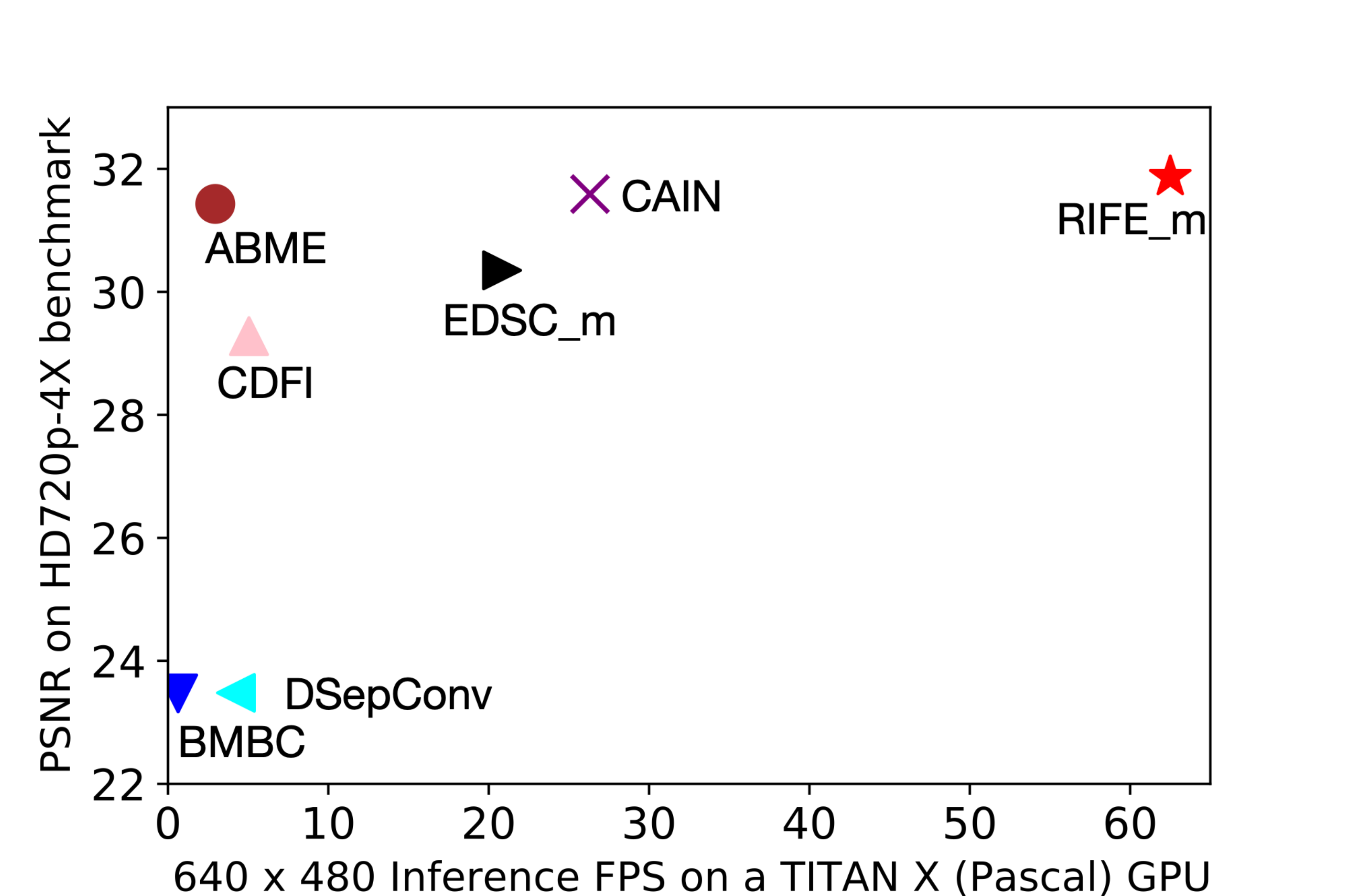}
	\vspace{-0.5em}
	\caption{\textbf{Performance comparison.} Results are reported for Vimeo90K~\cite{xue2019video} and HD-$4\times$~\cite{bao2019depth} benchmark. More details are in the experimental section} 
	\label{fig:intro_fig}
	\vspace{-1em}
\end{figure}

Combining these designs, we propose the Real-time Intermediate Flow Estimation (\textbf{RIFE}). RIFE trained from scratch can achieve satisfactory results, without requiring pre-trained models or datasets with optical flow labels. We illustrate the RIFE's performance compared with other methods in Figure~\ref{fig:intro_fig}. 

To sum up, our main contributions include:
\begin{itemize}
    \vspace{-0.5em}
	\item We design an effective IFNet to approximate the intermediate flows and introduce a privileged distillation scheme to improve the performance.
	\item Our experiments demonstrate that RIFE achieves SOTA performance on several public benchmarks, especially in the scene of arbitrary-time frame interpolation. 
	\item We show RIFE can be extended to applications such as depth map interpolation and dynamic scene stitching, thanks to its flexible temporal encoding.
\end{itemize}





	\section{Related Works}
\noindent\textbf{Optical Flow Estimation.} Optical flow estimation is a long-standing vision task that aims to estimate the per-pixel motion, useful in many downstream tasks~\cite{sun2018optical,zhou2022responsive,lu2019dvc,zhao2022tracking}. Since the milestone work of FlowNet~\cite{dosovitskiy2015flownet}, flow model architectures have evolved for several years, yielding more accurate results while being more efficient, such as FlowNet2.0~\cite{ilg2017flownet}, PWC-Net~\cite{sun2018pwc} and LiteFlowNet~\cite{hui2018liteflownet}. Recently Teed~\etal~\cite{teed2020raft} introduce RAFT, which iteratively updates a flow field through a recurrent unit and achieves a remarkable breakthrough in this field. Another important research direction is unsupervised optical flow estimation~\cite{meister2017unflow,jonschkowski2020matters,luo2020upflow} which tackles the difficulty of labeling.




\noindent\textbf{Video Frame Interpolation.} Recently, optical flow has been a prevalent component in video interpolation. In addition to the method of directly estimating the intermediate flow~\cite{liu2017video,lee2020adacof,park2020bmbc}, Jiang~\etal~\cite{jiang2018super} propose SuperSlomo using the linear combination of the two bi-directional flows as an initial approximation of the intermediate flows and then refining them using U-Net. Reda~\etal~\cite{reda2019unsupervised} and Liu~\etal~\cite{liu2019cyclicgen} propose to improve intermediate frames using cycle
consistency. Bao~\etal~\cite{bao2019depth} propose DAIN to estimate the intermediate flow as a weighted combination of bidirectional flow. Niklaus~\etal~\cite{niklaus2020softmax} propose SoftSplat to forward-warp frames and their feature map using softmax splatting. Xu~\etal~\cite{xu2019quadratic} propose QVI to exploit four consecutive frames and flow reversal filter to get the intermediate flows. Liu~\etal~\cite{liu2020enhanced} further extend QVI with rectified quadratic flow prediction to EQVI. 

Along with flow-based methods, flow-free methods have also achieved remarkable progress. Meyer~\etal~\cite{meyer2015phase} utilize phase information to learn the motion relationship for multiple video frame interpolation. Niklaus~\etal~\cite{Niklaus_ICCV_2017} formulate VFI as a spatially adaptive convolution whose convolution kernel is generated using a CNN given the input frames. Cheng~\etal propose DSepConv~\cite{cheng2020video} to extend kernel-based method using deformable separable convolution and. Choi~\etal~\cite{choi2020channel} propose an efficient flow-free method named CAIN, which employs the PixelShuffle operator and channel attention to capture the motion information implicitly. Some work further 
focus on increasing the resolution and frame rate of the video together and has achieved good visual effect~\cite{xiang2020zooming,xu2021temporal}. In addition, large-motion and animation frame interpolation is also fields of great interest~\cite{reda2022film,siyao2021deep,briedis2021neural}.

\begin{figure*}[tb]
	\centering
	\includegraphics[width=11cm]{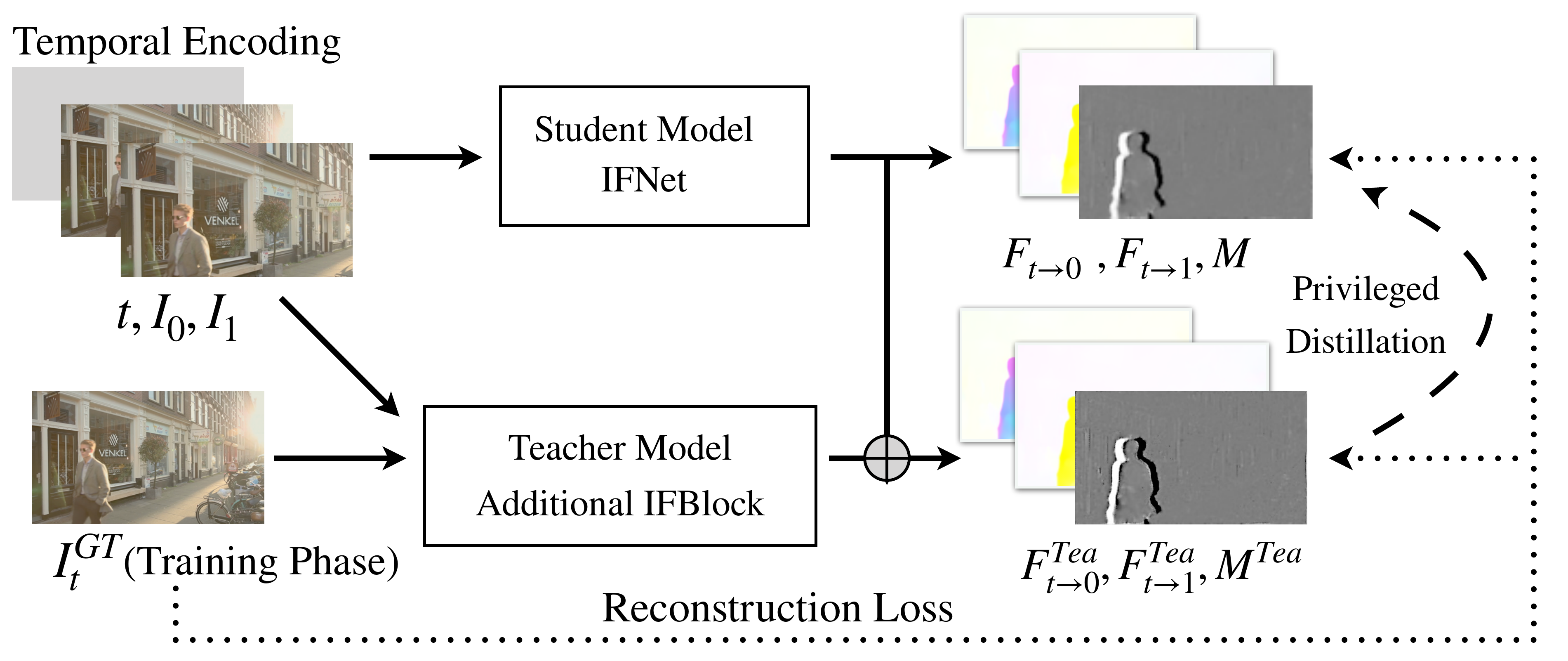}
	\vspace{-0.5em}
	\caption{\textbf{Overview of RIFE pipeline.} Given two input frames $I_0, I_1$ and temporal encoding $t$~(timestep encoded as an separate channel~\cite{mnih2013playing,huang2019learning}), we directly feed them into the IFNet to approximate intermediate flows $F_{t\rightarrow 0}, F_{t\rightarrow 1}$ and the fusion map $M$. During the training phase, a privileged teacher refines student's results based on ground truth $I_t$ using a special IFBlock }\label{fig:main}
\end{figure*}

\noindent\textbf{Knowledge Distillation.} Our privileged distillation~\cite{lopez2015unifying} for intermediate flow conceptually belongs to the knowledge distillation~\cite{hinton2015distilling}, which originally aims to transfer knowledge from a large model to a smaller one. In privileged distillation, the teacher model gets more input than the student model, such as scene depth, images from other views, and even image annotation. Therefore, the teacher model can provide more accurate representations to guide the student model to learn. This idea is applied to some computer vision tasks, such as hand pose estimation~\cite{yuan2018rgb}, re-identification~\cite{porrello2020robust} and video style transfer~\cite{chen2020optical}. Our work is also related to codistillation~\cite{anil2018large} where the student and teacher have the same architecture and different inputs during training.
	\section{Method}
\begin{figure}[t]
	\centering
	\includegraphics[width=10cm]{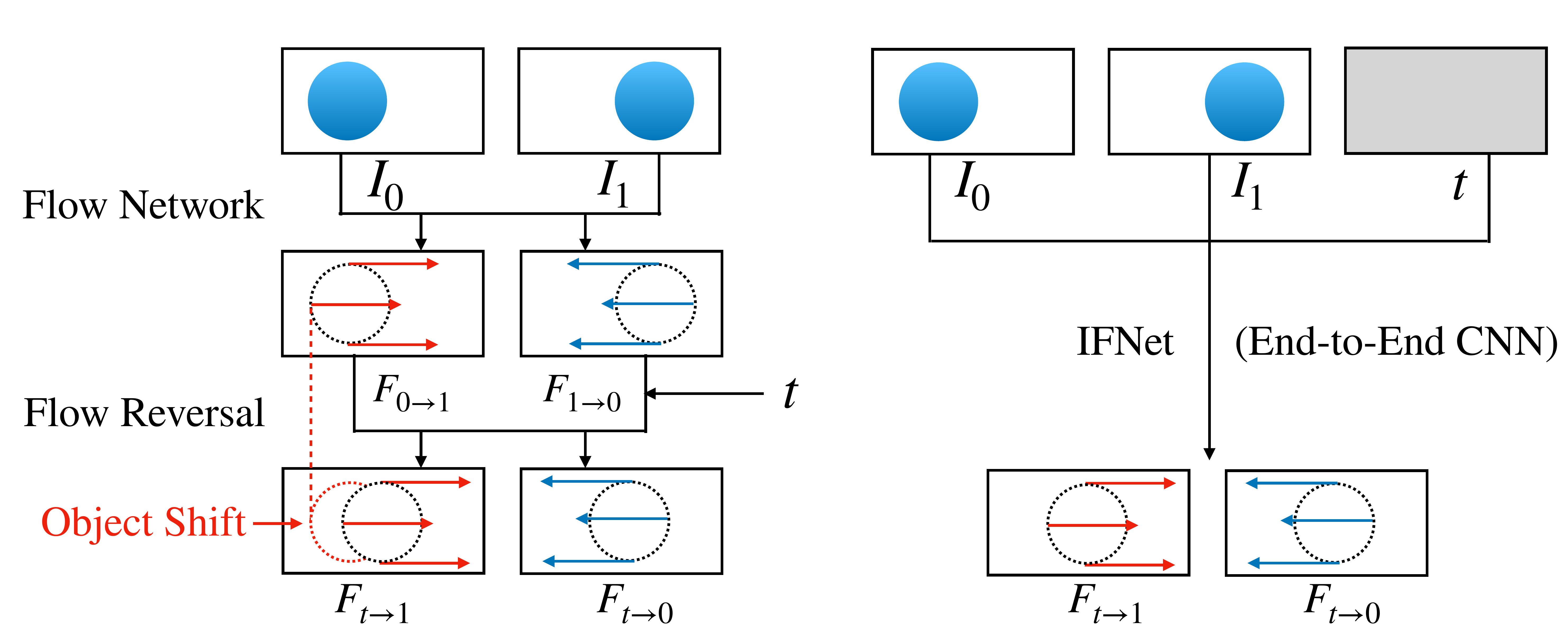}
	\vspace{-0.5em}
	\caption{\textbf{Compare indirect intermediate flow estimation~\cite{jiang2018super,xu2019quadratic,bao2019depth} (left) with IFNet (right).} As the object shifts, flow reversal modules may have flaws in motion boundaries. Rather than hand-engineering flow reversal layers, CNNs can learn intermediate flow estimates end-to-end}\label{fig:IFFlow}\vspace{-1em}
\end{figure}
\subsection{Pipeline Overview}
\label{subsec:overview}
We illustrate the overall pipeline of RIFE in Figure~\ref{fig:main}. Given a pair of consecutive RGB frames, $I_0, I_1$ and target timestep $t~(0 \leq t \leq 1)$, our goal is to synthesize an intermediate frame $\widehat{I}_t$. We estimate the intermediate flows $F_{t\rightarrow 0}$, $F_{t\rightarrow 1}$ and fusion map $M$ by feeding input frames and $t$ as an additional channel into the IFNet. We can get reconstructed image $\widehat{I}_t$ using following formulation:

\begin{equation}
\widehat{I}_t = M \odot \widehat{I}_{t\leftarrow 0} + (1 - M) \odot \widehat{I}_{t\leftarrow 1},
\end{equation}\vspace{-1em}
\begin{equation}
\widehat{I}_{t\leftarrow 0} = \backwardwarp(I_0, F_{t\rightarrow 0}),\quad \widehat{I}_{t\leftarrow 1} = \backwardwarp(I_1, F_{t\rightarrow 1}).
\end{equation}
where $\backwardwarp$ is the image backward warping, $\odot$ is an element-wise multiplier, and M is the fusion map $(0 \leq M \leq 1)$. We use another encoder-decoder CNNs named RefineNet following previous methods~\cite{jiang2018super,niklaus2020softmax} to refine the high-frequency area of $\widehat{I}_t$ and reduce artifacts of the student model. Its computational cost is similar to the IFNet. The RefineNet finally produce a reconstruction residual $\Delta~(-1\leq \Delta \leq 1)$. And we will get a refined reconstructed image $\widehat{I}_t + \Delta$. The detailed architecture of RefineNet is in the \textbf{Appendix}.

\subsection{Intermediate Flow Estimation}
\label{subsec:architecture}



Some previous VFI methods reverse and refine bi-directional flows~\cite{jiang2018super,xu2019quadratic,bao2019depth,liu2020enhanced} as depicted in Figure~\ref{fig:IFFlow}. The flow reversal process is usually cumbersome due to the difficulty of handling the changes of object positions. Intuitively, the previous flow reversal method hopes to perform spatial interpolation on the optical flow field, which is not trivial because of the ``object shift" problem. The role of our IFNet is to directly and efficiently predict $F_{t\rightarrow 0}, F_{t\rightarrow 1}$ and fusion mask $M$ given two consecutive input frames $I_0, I_1$ and timestep $t$. When $t=0$ or $t=1$, IFNet is similar to the classical optical flow models.

\begin{figure}[t]
	\centering
	\includegraphics[width=8.5cm]{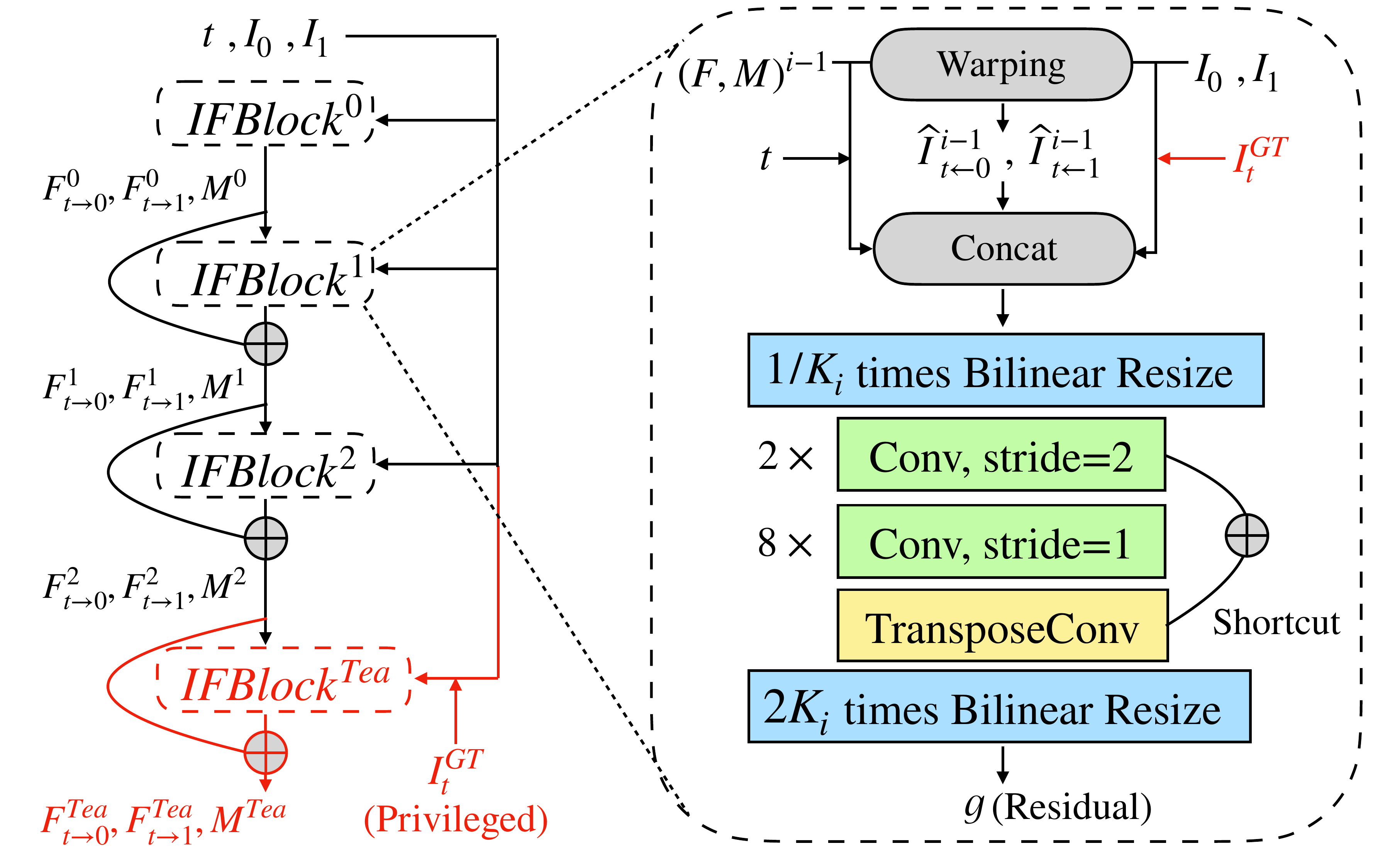}
	\vspace{-1em}
	\caption{\textbf{Left}: The IFNet is composed of several stacked IFBlocks operating at different resolution. \textbf{Right}: In an IFBlock, we first backward warp the two input frames based on current approximated flow $F^{i-1}$. Then the input frames $I_0, I_1$, warped frames $\widehat{I}_{t\leftarrow 0}, \widehat{I}_{t\leftarrow 1}$, the previous results $F^{i-1}, M^{i-1}$ and timestep $t$ are fed into the next IFBlock to approximate the residual of flow and mask. The privileged information $I_t^{GT}$ is only provided for teacher }\label{fig:IFNet}
\end{figure}

To handle the large motion encountered in intermediate flow estimation, we employ a coarse-to-fine strategy with gradually increasing resolution, as illustrated in Figure \ref{fig:IFNet}. Specifically, we first compute a rough prediction of the flow on low resolution, which is believed to capture large motions easier, then iteratively refine the flow fields with gradually increasing resolution. Following this design, our IFNet has a stacked hourglass structure, where a flow field is iteratively refined via successive IFBlocks:

\begin{equation}
\left[ \begin{array}{c}F^i \\ M^i \end{array} \right ]= \left[ \begin{array}{c}F^{i-1} \\ M^{i-1} \end{array} \right ]+ \text{IFB}^{i}(\left[ \begin{array}{c}F^{i-1} \\ M^{i-1} \end{array} \right ], t, \widehat{I}^{i-1}),
\end{equation}
where $F^{i-1}$ and $M^{i-1}$ denote the current estimation of the intermediate flows and fusion map from the $(i-1)^{th}$ IFBlock, and $\text{IFB}^{i}$ represents the $i^{th}$ IFBlock. We use a total of 3 IFBlocks, and each has a resolution parameter, $(K^0, K^1, K^2)=(4,2,1)$. During inference time, the final estimation is $F^n$ and $M^n(n=2)$. Each IFBlock has a feed-forward structure consisting of serveral convolutional layers and an up-sampling operator. Except for the layer that outputs the optical flow residuals and the fusion map, we use PReLU~\cite{he2015delving} as the activation function. The cost volume~\cite{dosovitskiy2015flownet} operator is computationally expensive and usually ties the starting point of optical flow to the input image. So it is not directly transferable.

\begin{table}[tb]
\vspace{-1em}
\caption{\textbf{Average inference time on the $640\times 480$ frames}. Recent VFI methods~\cite{jiang2018super,bao2019depth,niklaus2020softmax} run the optical flow model twice to obtain bi-directional flows}
\centering
\vspace{0.5em}
\begin{tabular}{ccccccc}
		\hline
		Method& FlowNet2.0~\cite{ilg2017flownet}&PWC-Net~\cite{sun2018pwc} &LiteFlownet~\cite{hui2018liteflownet}& RAFT~\cite{teed2020raft} & ~IFNet~\\ \hline
		Runtime &$2\times207$ms &$2\times21$ms &$2\times73$ms & $2\times52$ms& \first{7ms} \\ \hline
\end{tabular}
\label{tab:runtime}
\vspace{-1em}
\end{table}



We compare the runtime of the SOTA optical flow models~\cite{sun2018pwc,hui2018liteflownet,teed2020raft} and IFNet in Table~\ref{tab:runtime}. Current flow-based VFI methods~\cite{jiang2018super,bao2019depth,niklaus2020softmax} usually need to run their flow models twice then process the bi-directional flows. Therefore the intermediate flow estimation in RIFE runs at a faster speed. Although these optical models can estimate inter-frame motion accurately, they are not suitable for direct migration to VFI tasks.

\begin{figure}[t]
	\centering
	\includegraphics[width=12cm]{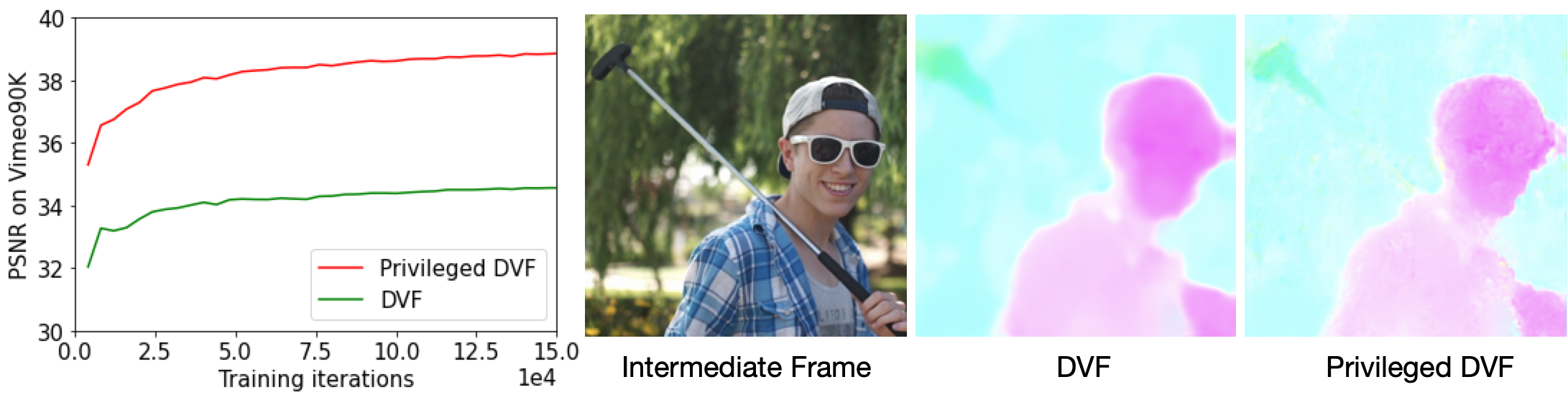}
	\caption{\textbf{Results of DVF~\cite{liu2017video}~(Vimeo90K)}. After feeding the edge map of intermediate frames~(privileged information) into the model, the estimated flows can be significantly improved, resulting in better reconstruction on validation set} \label{leak}
\vspace{-1em}
\end{figure}

\subsection{Priveleged Distillation for Intermediate Flow}

\label{subsec:leakage}

We use an experiment to show that directly approximating the intermediate flows is challenging without access to the intermediate frame. We train DVF~\cite{liu2017video} model to estimate intermediate flow on Vimeo90K~\cite{xue2019video} dataset. As a comparison, we add an additional input channel to the DVF model, containing the edge map~\cite{ding2001canny} of intermediate frames~(denoted as ``Privileged DVF"). Figure~\ref{leak} shows that the quantization result of Privileged DVF is surprisingly high, while the flows estimated by DVF are blurry. Similar conclusions are also demonstrated in deferred rendering, showing that VFI will be simpler with some intermediate information~\cite{briedis2021neural}. This demonstrates that estimating optical flow between two images is easier for the model than estimating intermediate flow. This inspire us to design a privileged model to teach the original model.

We design a privileged distillation loss to IFNet. We stack an additional IFBlock (teacher model $\text{IFB}^{Tea}$, $K^{Tea} = 1$) that refines the results of IFNet referring to the target frame $I^{GT}_t$: 

\begin{equation}
\left[ \begin{array}{c}F^{Tea} \\ M^{Tea} \end{array} \right ] = \left[ \begin{array}{c}F^n \\ M^n \end{array} \right ] + \text{IFB}^{Tea}(\left[ \begin{array}{c}F^n \\ M^n \end{array} \right ], t, \widehat{I}^n, I^{GT}_t).
\end{equation}

With the access of $I^{GT}_t$ as privileged information, the teacher model produces more accurate flows. We define the distillation loss $\mathcal{L}_{dis}$ as follows:
\begin{equation}
\mathcal{L}_{dis} = \sum_{i\in\{0, 1\}}|| F_{t\rightarrow i} - F^{Tea}_{t\rightarrow i}||_2.
\end{equation} 
We apply the distillation loss over the full sequence of predictions generated from the iteratively updating process in the student model. The gradient of this loss will not be backpropagated to the teacher model. The teacher block will be discarded after the training phase, hence this would incur no extra cost for inference. It makes more stable training and faster convergence.








\subsection{Implementation Details}
\label{subsec:implement}

\noindent\textbf{Supervisions.} Our training loss $\mathcal{L}_{total}$ is a linear combination of the reconstruction losses $\mathcal{L}_{rec}, \mathcal{L}^{Tea}_{rec}$ and privileged distillation loss $\mathcal{L}_{dis}$:  
\begin{equation}
\mathcal{L}_{total} = \mathcal{L}_{rec} + \mathcal{L}^{Tea}_{rec} + \lambda_{d} \mathcal{L}_{dis}
\label{eq:total_loss},
\end{equation}
where we set $\lambda_{d} = 0.01$ to balance the scale of losses.

The reconstruction loss $\mathcal{L}_{rec}$ models the reconstruction quality of the intermediate frame. The reconstruction loss has the formulation of:
\begin{equation}
\mathcal{L}_{rec} = d(\widehat{I}_t, {I}^{GT}_{t}), \mathcal{L}_{rec}^{Tea} = d(\widehat{I}_t^{Tea},  {I}^{GT}_{t}) 
\label{eq:loss},
\end{equation}
where $d$ is often a pixel-wised loss. Following previous work~\cite{niklaus2018context,niklaus2020softmax}, we use $L_1$ loss between two Laplacian pyramid representations of the reconstructed image and ground truth (denoted as $L_{Lap}$, the pyramidal level is $5$).
%
%


\noindent\textbf{Training Dataset.}
We use the Vimeo90K dataset~\cite{xue2019video} to train RIFE. This dataset has $51,312$ triplets for training, where each triplet contains three consecutive video frames with a resolution of $448\times256$. We randomly augment the training data using horizontal and vertical flipping, temporal order reversing, and rotating by $90$ degrees.

\noindent\textbf{Training Strategy.}
We train RIFE on the Vimeo90K training set and fix $t=0.5$. RIFE is optimized by AdamW~\cite{loshchilov2018fixing} with weight decay $10^{-4}$ on $224\times224$ patches. Our training uses a batch size of $64$. We gradually reduce the learning rate from $10^{-4}$ to $10^{-5}$ using cosine annealing during the whole training process. We train RIFE on $8$ TITAN X (Pascal) GPUs for $300$ epochs in $10$ hours. 

We use the Vimeo90K-Septuplet~\cite{xue2019video} dataset to extend RIFE to support arbitrary-timestep frame interpolation~\cite{cheng2020multiple,kalluri2020flavr}. This dataset has $91,701$ sequence with a resolution of $448\times256$, each of which contains $7$ consecutive frames. For each training sample, we randomly select $3$ frames $(I_{n_0}, I_{n_1}, I_{n_2})$ and calculate the target timestep $t = (n_1 - n_0) / (n_2 - n_0)$, where $0\leq n_0 < n_1 < n_2 < 7$. So we can write RIFE's temporal encoding to extend it. We keep other training setting unchanged and denote the model trained on Vimeo90K-Septuplet as RIFE$_m$. 
	\section{Experiments}

\begin{figure}[t]
	\centering
	\includegraphics[width=12cm]{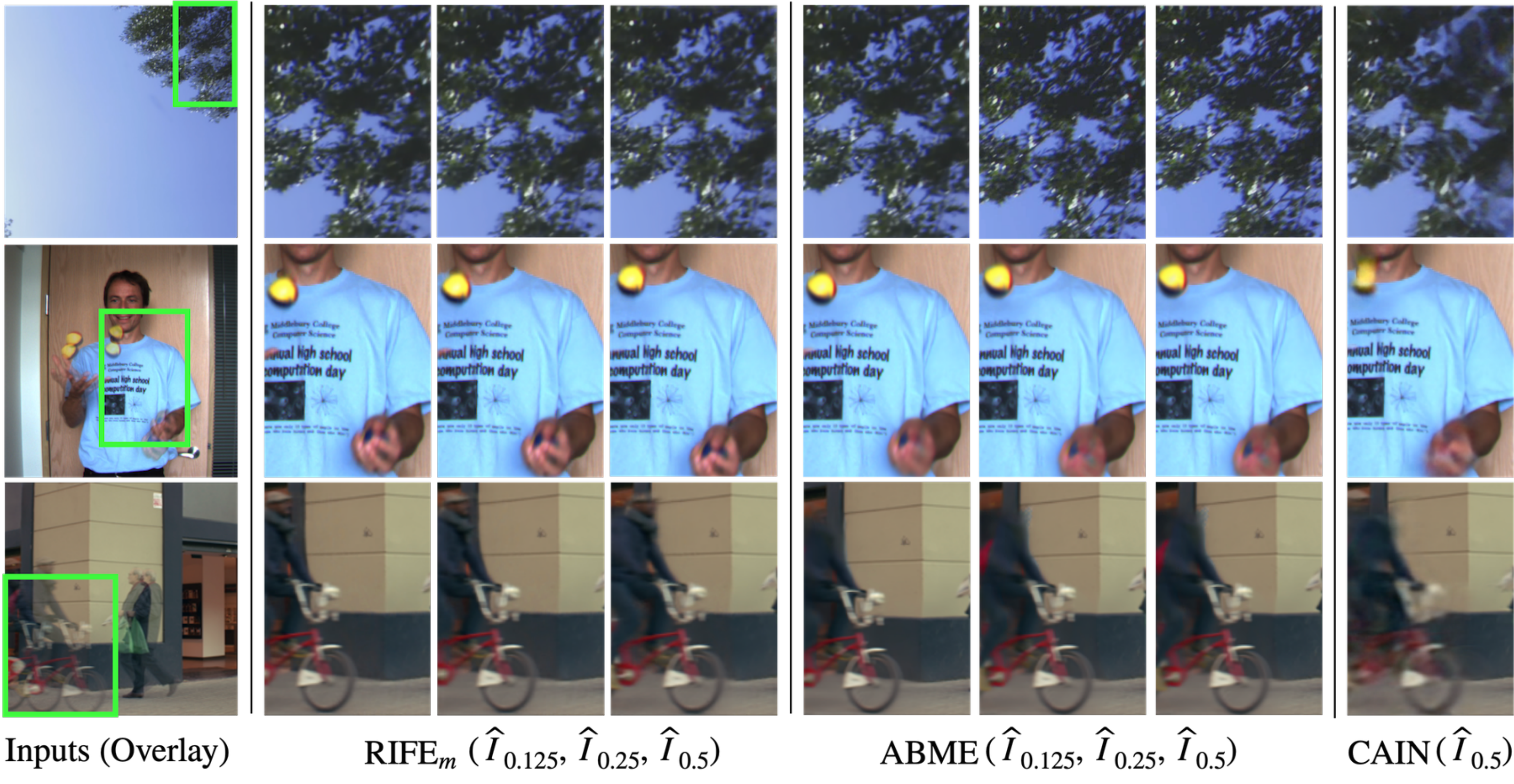}
	\vspace{-0.5em}
	\caption{\textbf{Interpolating multiple frames using RIFE$_m$.} These images are from HD~\cite{bao2019depth}, M.B.~\cite{baker2011database}, Vimeo90K~\cite{xue2019video} benchmarks, respectively. We attach the results of CAIN~\cite{choi2020channel} and ABME~\cite{park2021ABME}. RIFE$_m$ provides smooth and continuous motions}\label{fig:multi}
	\vspace{-1em}
\end{figure}

We first introduce the benchmarks for evaluation. Then we provide variants of our models with different computational costs. We compare these models with representative SOTA methods. In addition, we show the capability of generating arbitrary-timestep frames and other applications using RIFE. An ablation study is carried out to analyze our design. Finally, we discuss some limitations of RIFE.

\subsection{Benchmarks and Evaluation Metrics}
We train our models on the Vimeo90K training dataset and directly test it on the following benchmarks. 


\noindent\textbf{Vimeo90K}. There are 3,782 triplets in the Vimeo90K testing set~\cite{xue2019video} with resolution of $448 \times 256$. This dataset is widely evaluated in recent VFI methods.
%

\noindent\textbf{UCF101.}
The UCF101 dataset~\cite{soomro2012ucf101} contains videos with various human actions. There are 379 triplets with a resolution of $256 \times 256$.


\noindent\textbf{HD.}
Bao~\etal~\cite{bao2019memc} collect 11 videos for evaluation. The HD benchmark consists of four 1080p, three 720p and four $1280 \times 544$ videos. Following the author of this benchmark, we use the first 100 frames of each video for evaluation. 

\noindent\textbf{X4K-1000FPS.} A recently released high frame rate 4K dataset~\cite{sim2021xvfi} containing $15$ scenes for testing. We follow the evaluation of~\cite{park2021ABME,park2021ABME}.

We measure the peak signal-to-noise ratio (PSNR), structural similarity (SSIM), and interpolation error (IE) for quantitative evaluation. All the methods are tested on a TITAN X (Pascal) GPU. To report the runtime, we test all models for processing a pair of $640\times 480$ images using the same device. Disagreements with some of the published results are explained in the \textbf{Appendix}.

\subsection{Comparisons with Previous Methods}
\label{sec:comparison}

We compare RIFE with previous VFI models  \cite{xue2019video,Niklaus_ICCV_2017,bao2019memc,bao2019depth,choi2020channel,niklaus2020softmax,park2020bmbc,cheng2020video,lee2020adacof,ding2021cdfi,cheng2020multiple}. These models are officially released except SoftSplat~\cite{niklaus2020softmax}. A recently unofficial reproduction~\cite{reda2022film} report SoftSplat~\cite{niklaus2020softmax} is slower than ABME~\cite{park2021ABME}, and we can not verify it with the available materials. In addition, we train DVF~\cite{liu2017video} model and SuperSlomo~\cite{jiang2018super} using our training pipeline on Vimeo90K dataset because the released models of these methods are trained on early datasets. 

\begin{table}[t]
	\caption{\textbf{Quantitative evaluation~(PSNR) for $4\times$ interpolation on the HD~\cite{bao2019memc} and 8$\times$ interpolation on X4K-1000FPS~\cite{sim2021xvfi} benchmark. } The 544p videos of HD benchmark are relatively more dynamic. Thus the PSNR index of these 544p videos is lower}
	\centering
	\vspace{1.5mm}
	\small
	\begin{tabular}{lcccccc}
		\hline
		Method & Arbitrary-timestep~~ & HD544p~ & HD720p~ & HD1080p~ & X4K-1000FPS\\ \hline
		DAIN~\cite{bao2019depth} & \checkmark& 22.17 & 30.25 & OOM & 26.78$\dag$ \\
		CAIN~\cite{choi2020channel} & - & 21.81 & 31.59 & 31.08 & - \\
		BMBC~\cite{park2020bmbc} & \checkmark & 19.51 & 23.47 & OOM & OOM
		\\ 
		DSepconv~\cite{cheng2020video} & - & 19.28 & 23.48 & OOM & OOM\\
		CDFI~\cite{ding2021cdfi} & - & 21.85 & 29.28 & OOM & OOM\\
		EDSC$_m$~\cite{cheng2020multiple} & \checkmark & 21.89 & 30.35 & 30.91 & -\\
		ABME~\cite{park2021ABME} & - & 22.46 & 31.43 & 33.22 & 30.16$\dag$\\
		\hline
		RIFE$_m$ & \checkmark & \textbf{22.95} & \textbf{31.87} & \textbf{34.25} & \textbf{30.58}$\ddag$ \\ \hline
	\end{tabular}
	\normalsize
	
	\begin{tablenotes}
		\raggedleft
		\item{
   
    $\dag$: copy from~\cite{park2021ABME}.
    $\ddag$: estimate flows on 1/4 downsampled videos.
	}
	\end{tablenotes}
	\label{tab:hd8}
\end{table}

\vspace{0.5em}
\noindent\textbf{Interpolating Arbitrary-timestep Frame.}
\label{sec:model_multi}
Arbitrary-timestep VFI is important in frame-rate conversion. We apply RIFE$_m$ to interpolate multiple intermediate frames at different timesteps $t \in (0, 1)$, as shown in Figure~\ref{fig:multi}. RIFE$_m$ can successfully handle $t=0.125~(8\times)$ which is not included in the training data.

To provide a quantitative comparison of multiple frame interpolation, we further extract every fourth frame of videos from HD benchmark~\cite{bao2019memc} and use them to interpolate other frames. We divide the HD benchmark into three subsets with different resolution to test these methods. We show the quantitative PSNR between generated frames and frames of the original videos in Table~\ref{tab:hd8}. Note that DAIN~\cite{bao2019depth}, BMBC~\cite{park2020bmbc} and EDSC$_m$~\cite{cheng2020video} can generate a frame at an arbitrary timestep. Some other methods can only interpolate the intermediate frame at $t=0.5$. Thus we use them recursively to produce $4\times$ results. Specifically, we firstly apply the single interpolation method once to get intermediate frame $\widehat I_{0.5}$. Then we feed $I_0$ and $\widehat I_{0.5}$ to get $\widehat I_{0.25}$ and so on. Furthermore, we test $8\times$ interpolation in a recently released dataset, X4K-1000FPS~\cite{sim2021xvfi}. Overall, RIFE$_m$ is very effective in the multiple frame interpolation. 

\noindent\textbf{Model Scaling.} \label{sec:model_scaling}To scale our models that can be compared with existing methods, we introduce two modifications following: test-time augmentation and resolution multiplying. 1) We flip the input images horizontally and vertically to get augmented test data. We infer and average~(with flipping) these two results finally. This model is denoted as RIFE-{2T}. 2) We remove the first downsample layer of IFNet and add a downsample layer before its output to match the origin pipeline. We also perform this modification on RefineNet. It enlarges the process resolution of the feature maps and produces a model named RIFE-{2R}. We combine these two modifications to extend RIFE to RIFE-Large ({2T2R}).


\definecolor{mygray}{gray}{.9}
\begin{table*}[tb]
\caption{
		\textbf{Quantitative comparisons on several benchmarks.} The images of each dataset are directly inputted to each model. Some models are unable to run on 1080p images due to exceeding the  memory available on our graphics card (denoted as ``OOM”). We use gray backgrounds to mark the methods that require pre-trained depth models or optical flow models
	}
	\vspace{1mm}
	\centering
	\resizebox{1\textwidth}{!}{\begin{tabular}{lcccccccccc}
		\toprule
		\multirow{2}{*}[-0.28em]{Method}  &\# Parameters&
		Runtime
&\multicolumn{2}{c}{UCF101~\cite{soomro2012ucf101}} &\multicolumn{2}{c}{Vimeo90K~\cite{xue2019video}} & M.B.~\cite{baker2011database} & ~~HD~\cite{bao2019depth}~~\\

		\cmidrule(l{7pt}r{7pt}){4-5}
		\cmidrule(l{7pt}r{7pt}){6-7}
		\cmidrule(l{5pt}r{5pt}){8-8}
		\cmidrule(l{5pt}r{5pt}){9-9}
		\vspace{0.2em}
		&(Million) & (ms) &PSNR & SSIM 	&PSNR & SSIM & IE & PSNR\\
		\addlinespace[-1pt]
		\midrule
		
		
			DVF~\cite{xue2019video} &1.6 &80 & {34.92} & {0.968} &{34.56}& {0.973}  &{2.47} & 31.47 \\
		
		TOFlow~\cite{baker2011database} &~\textbf{1.1} &84 & 34.58 & 0.967  &33.73& 0.968 &2.15 &29.37  \\
		
		
		 \rowcolor{mygray}DAIN~\cite{bao2019depth} & 24.0 & 436 & {35.00} & {0.968} &{34.71}& {0.976}  &2.04 & 31.64$^{\dag}$\\ 
		
		DSepConv~\cite{cheng2020video} & 21.8 & 236 & 35.08 & 0.969 & 34.73 & 0.974 & 2.03 & OOM\\
		
		\rowcolor{mygray} SoftSplat~\cite{niklaus2020softmax}$^{\dag}$ & ~7.7 & - & 35.39 & \textbf{0.970} & 36.10 & 0.980 & \textbf{1.81} & - \\

		BMBC~\cite{park2020bmbc} & 11.0 & 1580 & 35.15 & {0.969} & 35.01 & 0.976 & 2.04 & OOM\\
		
		CDFI~\cite{ding2021cdfi} & ~5.0 & 198 & 35.21 & 0.969 &35.17 & 0.977 & 1.98 & OOM\\
		
		ABME~\cite{park2021ABME} & 18.1 & 339 & 35.37 & \textbf{0.970} & 36.18 & \textbf{0.981} & 1.88 & 32.17
		
		\\
		
		RIFE-Large & {~9.8} & 80 & \textbf{35.41} & \textbf{0.970} & \textbf{36.19} & \textbf{0.981} & 1.82 & \textbf{32.31}\\
		
		\hline
		\textit{Relatively Fast Models} \\
		CAIN~\cite{choi2020channel} & 42.8 & 38 & 34.98 & \textbf{0.969} & 34.65 & 0.973 & 2.28 & 31.77\\
		
		Superslomo~\cite{jiang2018super} &19.8 &62 & {35.15} & {0.968} &{34.64}& {0.974}  &{2.21} & 31.55\\
		
		SepConv~\cite{niklaus2017video} & 21.6 & 51 & 34.78 & 0.967  &33.79& {0.970} &2.27& 30.87 \\
		
		AdaCoF~\cite{lee2020adacof} & 21.8 & 34 & 34.91 & 0.968 & 34.27 & 0.971 & 2.31 & 31.43 \\ 
		
		EDSC~\cite{cheng2020multiple} & \textbf{~8.9} & 46 & 35.13 & 0.968 & 34.84 & 0.975 & 2.02 & 31.59 \\
		
		RIFE & {~9.8} & \textbf{16} & \textbf{35.28} & \textbf{0.969} & \textbf{35.61} & \textbf{0.978} & \textbf{1.96} & 32.14\\
		
		RIFE$_m\ddag$ & {~9.8} & \textbf{16} & 35.22 & \textbf{0.969} & 35.46 & \textbf{0.978} & 2.16 & \textbf{32.31}\\
		\bottomrule
	\end{tabular}
	}
	\label{tab:comparison}
	\begin{tablenotes}
		\raggedleft
		\item{
    $\dag$: copy from the original papers. $\ddag$: trained on Vimeo90K-Septuplet dataset.
	}
	\end{tablenotes}
	\label{tab:UCF101_Vimeo90K_MB}
	\vspace{-1em}
\end{table*}
\begin{figure*}[t]
  \centering
  \begin{minipage}[t]{0.145\linewidth}
  \centering
  \includegraphics[width=1\linewidth]{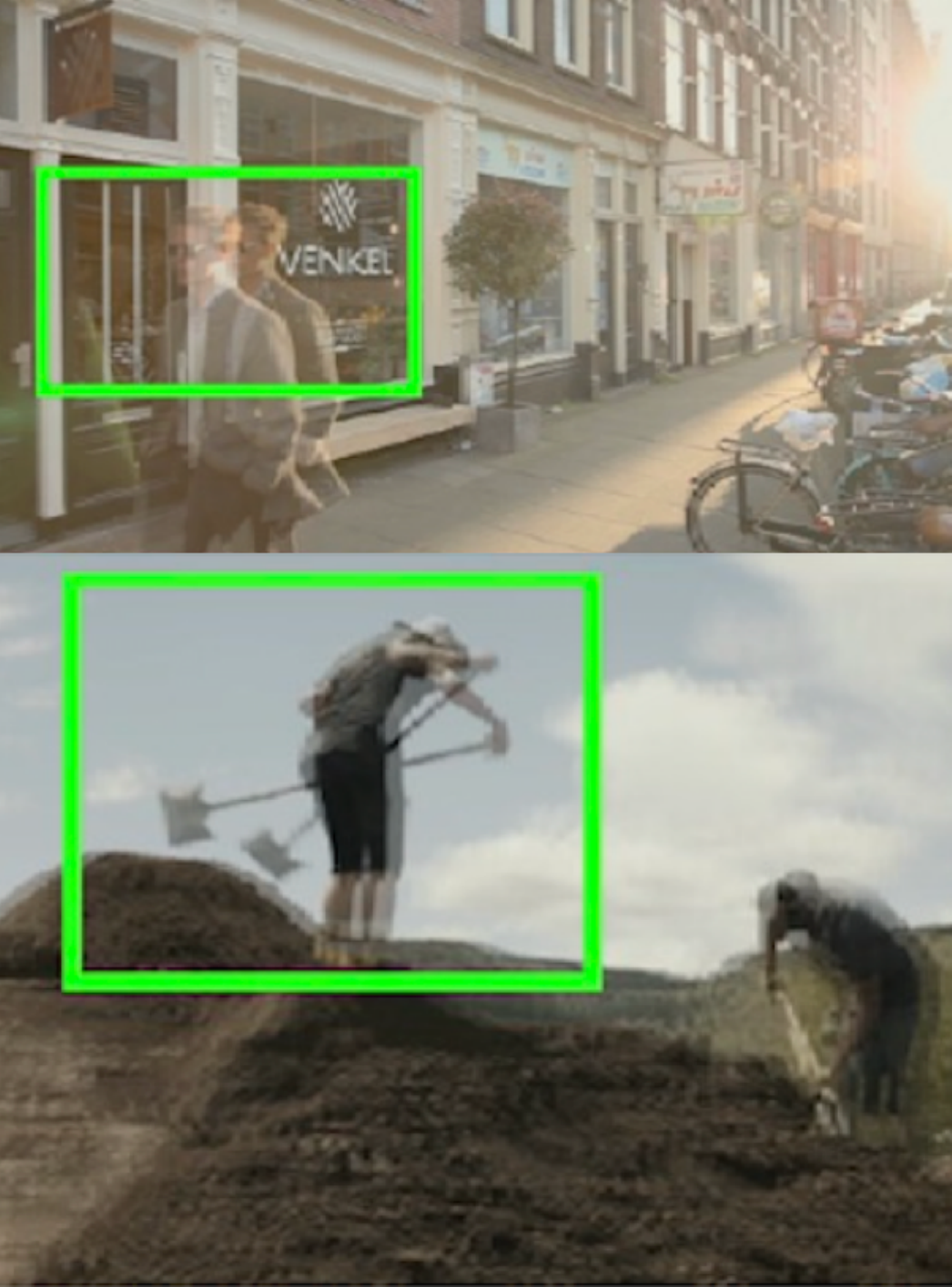}
  \centerline{Inputs}
  \end{minipage}
  \begin{minipage}[t]{0.145\linewidth}
  \centering
  \includegraphics[width=1\linewidth]{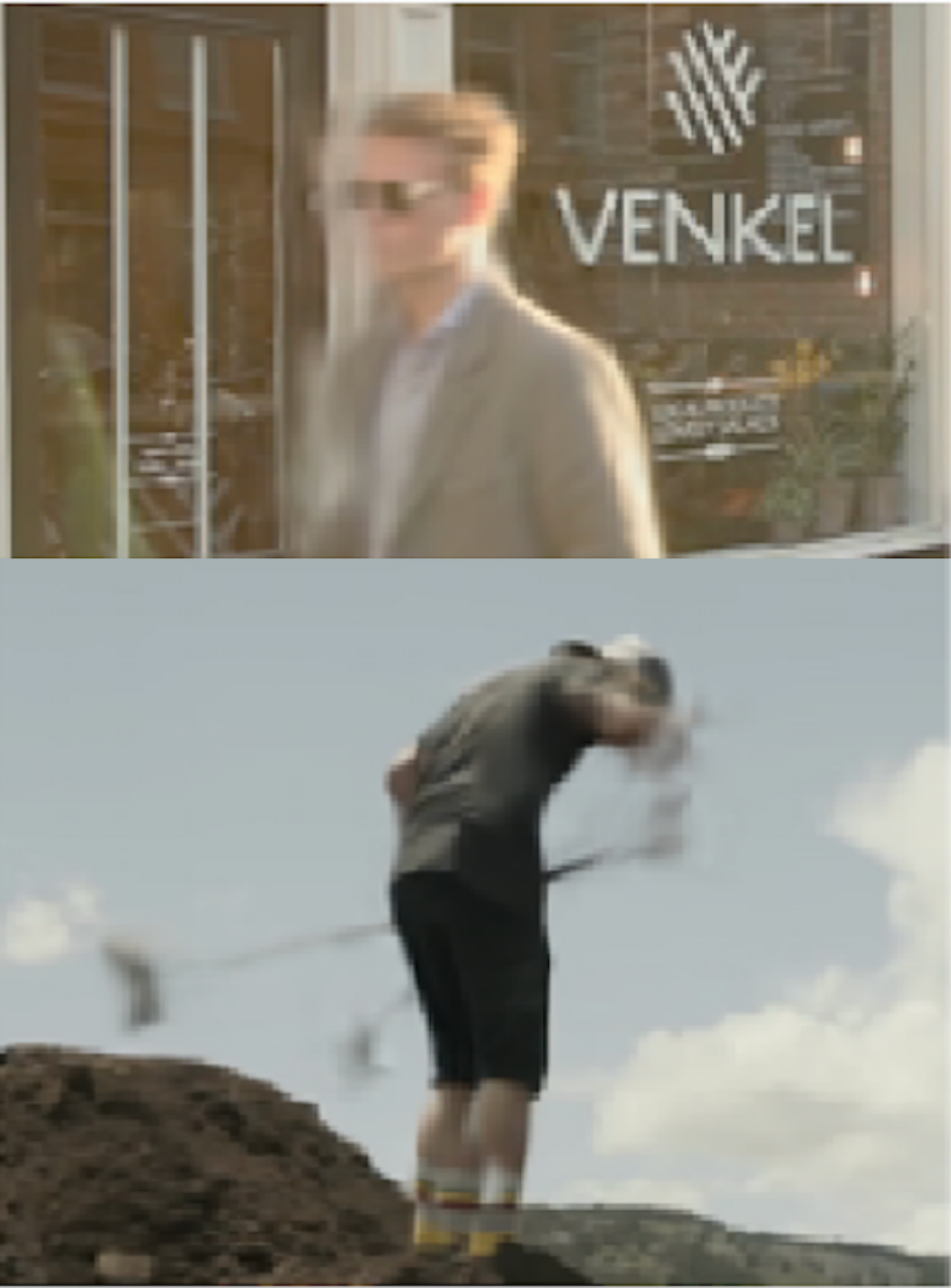}
  \centerline{SepConv~\cite{niklaus2017video}}
  \end{minipage}
  \begin{minipage}[t]{0.145\linewidth}
  \centering
  \includegraphics[width=1\linewidth]{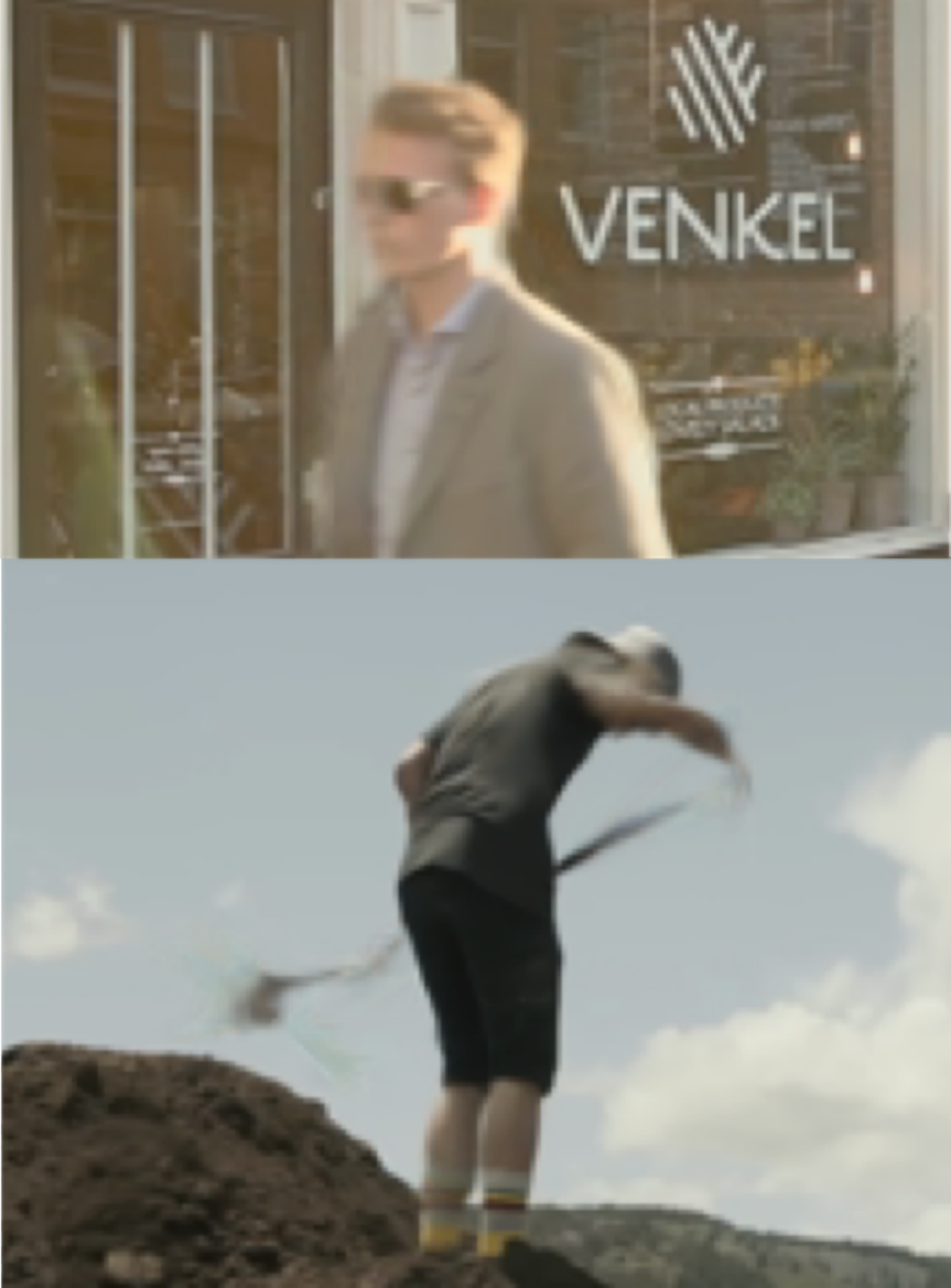}
  \centerline{DAIN~\cite{bao2019depth}}
  \end{minipage}
  \begin{minipage}[t]{0.145\linewidth}
  \centering
  \includegraphics[width=1\linewidth]{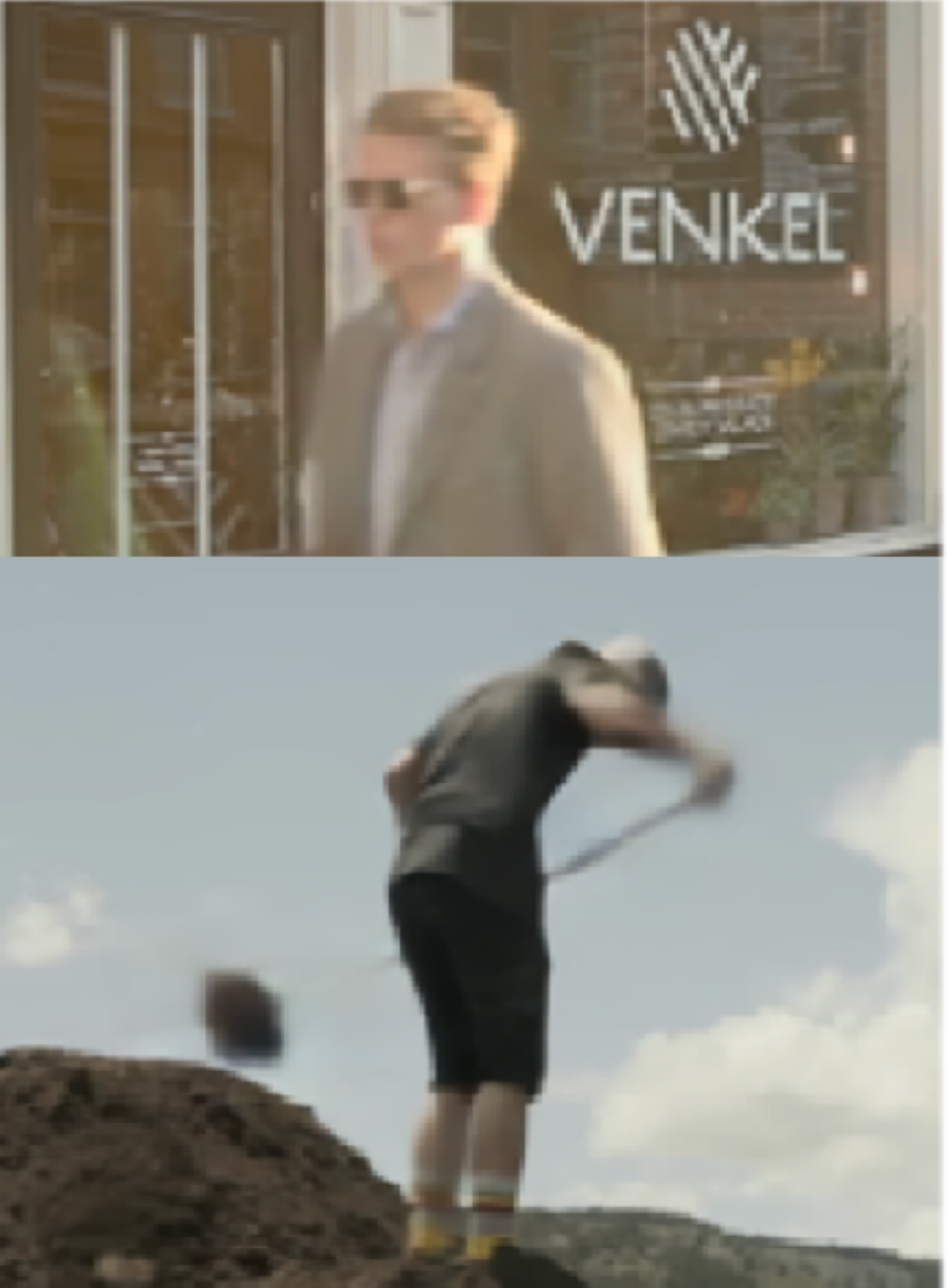}
  \centerline{CAIN~\cite{choi2020channel}}
  \end{minipage}
  \begin{minipage}[t]{0.146\textwidth}
  \centering
  \includegraphics[width=1\linewidth]{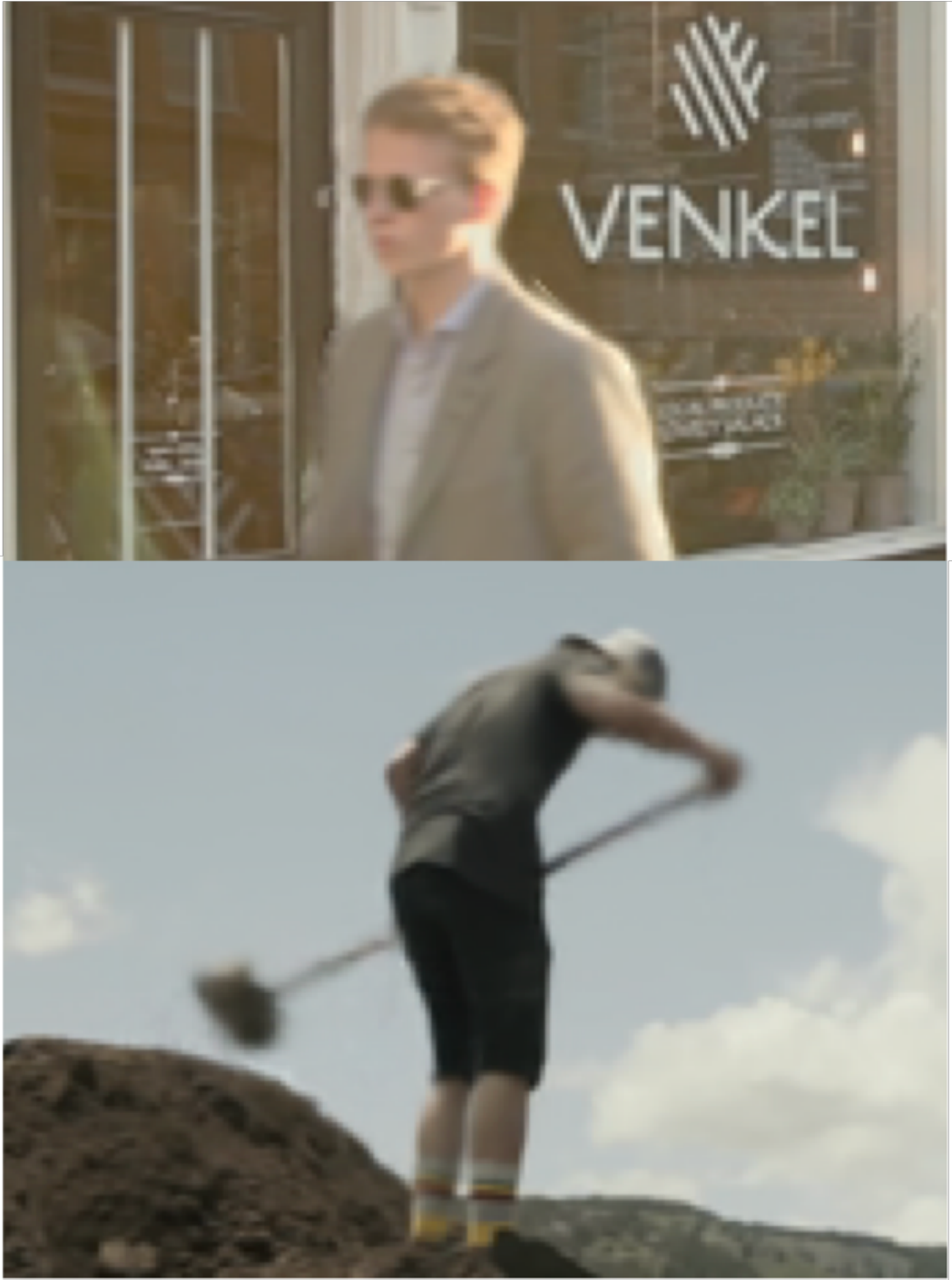}
  \centerline{RIFE (Ours)}
  \end{minipage}
  \begin{minipage}[t]{0.1465\textwidth}
  \centering
  \includegraphics[width=1\linewidth]{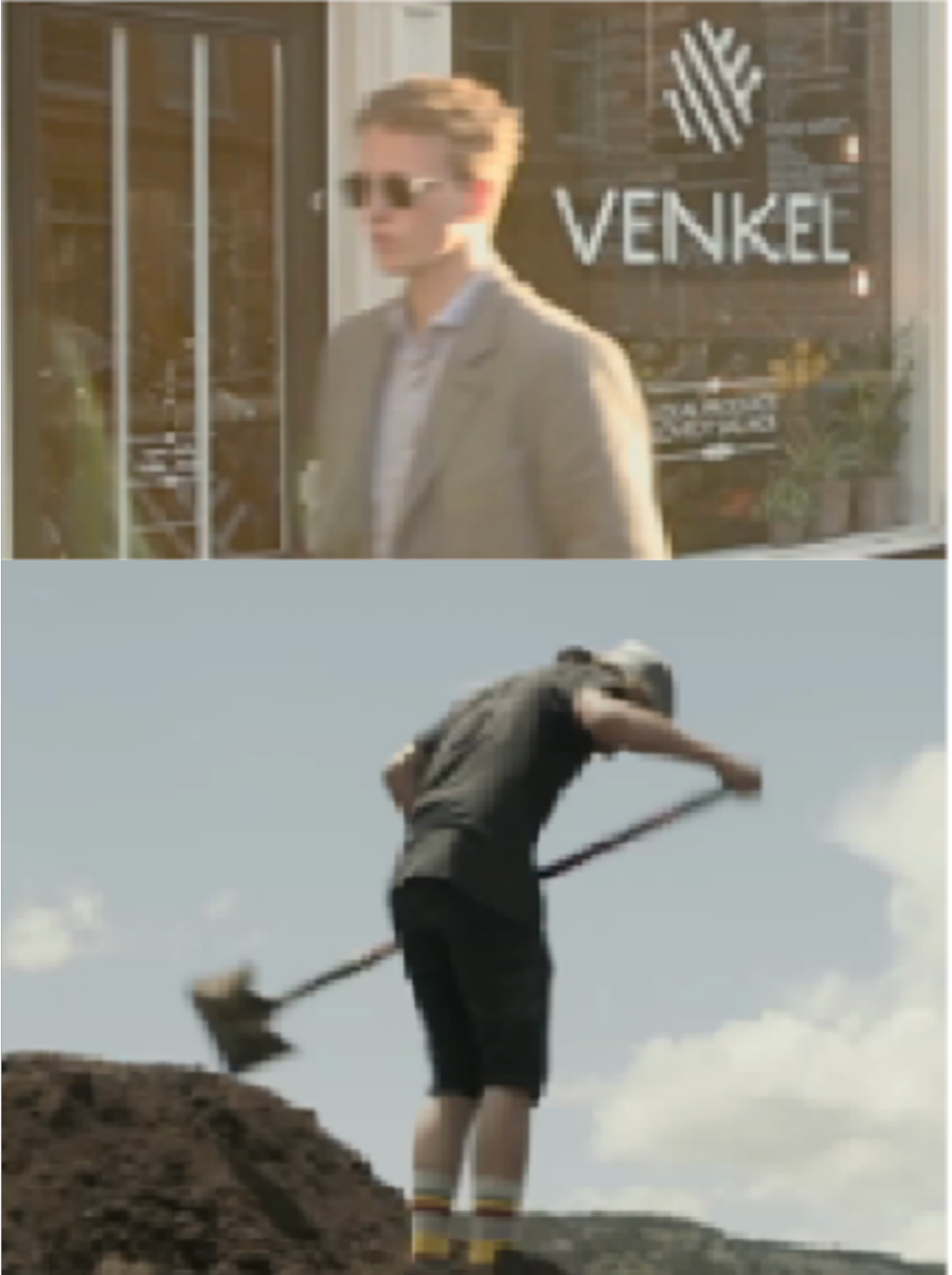}
  \centerline{GT}
  \end{minipage}
  \vspace{-0.5em}
  \caption{\textbf{Qualitative comparison on Vimeo90K~\cite{xue2019video} testing set} }
  \vspace{-0.5em}
\label{fig:Vimeo}
\end{figure*}

\noindent\textbf{Middle Timestep Interpolation}. We report the performance of middle timestep interpolation in Table~\ref{tab:UCF101_Vimeo90K_MB}. For ease of comparison, we group the models by running speed. RIFE achieve very high performance compared to other small models. Meanwhile, RIFE needs only about $3$ gigabytes of GPU memory to process 1080p videos. We get a larger version of our model~(RIFE-Large) by model scaling, which runs about $4\times$ faster than ABME~\cite{park2021ABME} with comparable performance. We provide a visual comparison of video clips with large motions from the Vimeo90K testing set in Figure~\ref{fig:Vimeo}, where SepConv~\cite{Niklaus_ICCV_2017} and DAIN~\cite{bao2019depth} produce ghosting artifacts, and CAIN~\cite{choi2020channel} causes missing-parts artifacts. Overall, RIFE~(with small computation) can produce more reliable results.

\begin{figure}[t]
	\centering
	\includegraphics[width=7cm]{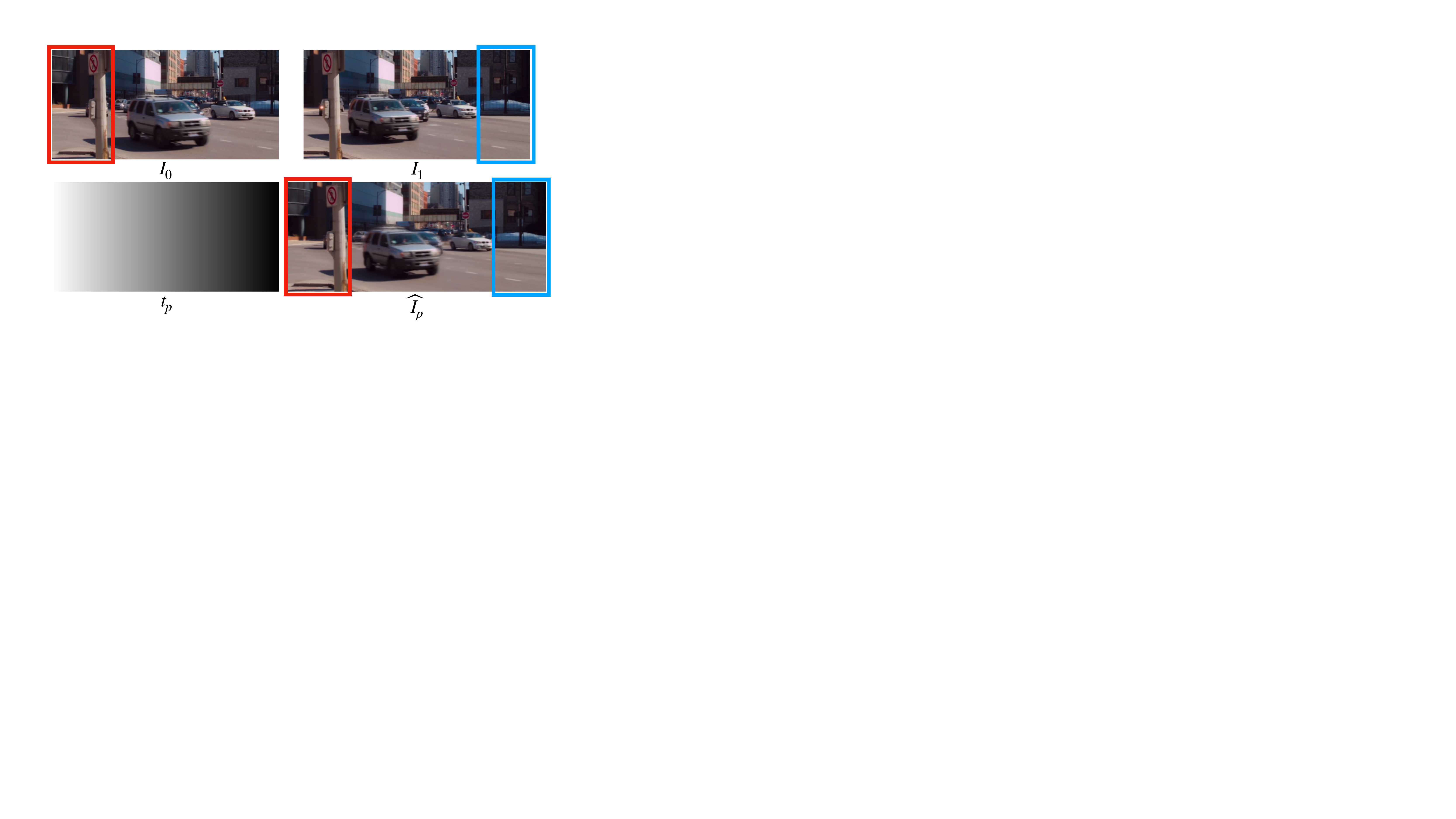}
	\vspace{-0.5em}
	\caption{\textbf{Synthesize images from two views on one ``panoramic" image $\widehat{I}_p$ using RIFE$_m$.} $\widehat{I}_p$ has been stretched for better visualization}\label{fig:Stitching} 
	\vspace{-1em}
\end{figure}

\subsection{General Temporal Encode}
In the VFI task, our temporal encoding $t$ is used to control
the timestep. To show its generalization capability, we demonstrate that we can control this encoding to implement diverse applications. As shown in Figure~\ref{fig:Stitching}, if we input a gradient encoding $t_p$, the RIFE$_m$ will synthesize the two images from dynamic scenes in a ``panoramic" view~ (use different timestamps for each column). The position relation of the vehicle in $\widehat{I}_p$ is between $I_0$ and $I_1$. In other words, if $I_0$, $I_1$ are from the binocular camera, the shooting time of $I_1$ is later than that of $I_0$. $\widehat{I}_p$ is the result of a wider FOV camera scan in columns. Similarly, this method may potentially eliminate the rolling shutter of the videos by having different timestamps for each horizontal row.

\begin{figure}[t]
	\centering
	\includegraphics[width=9cm]{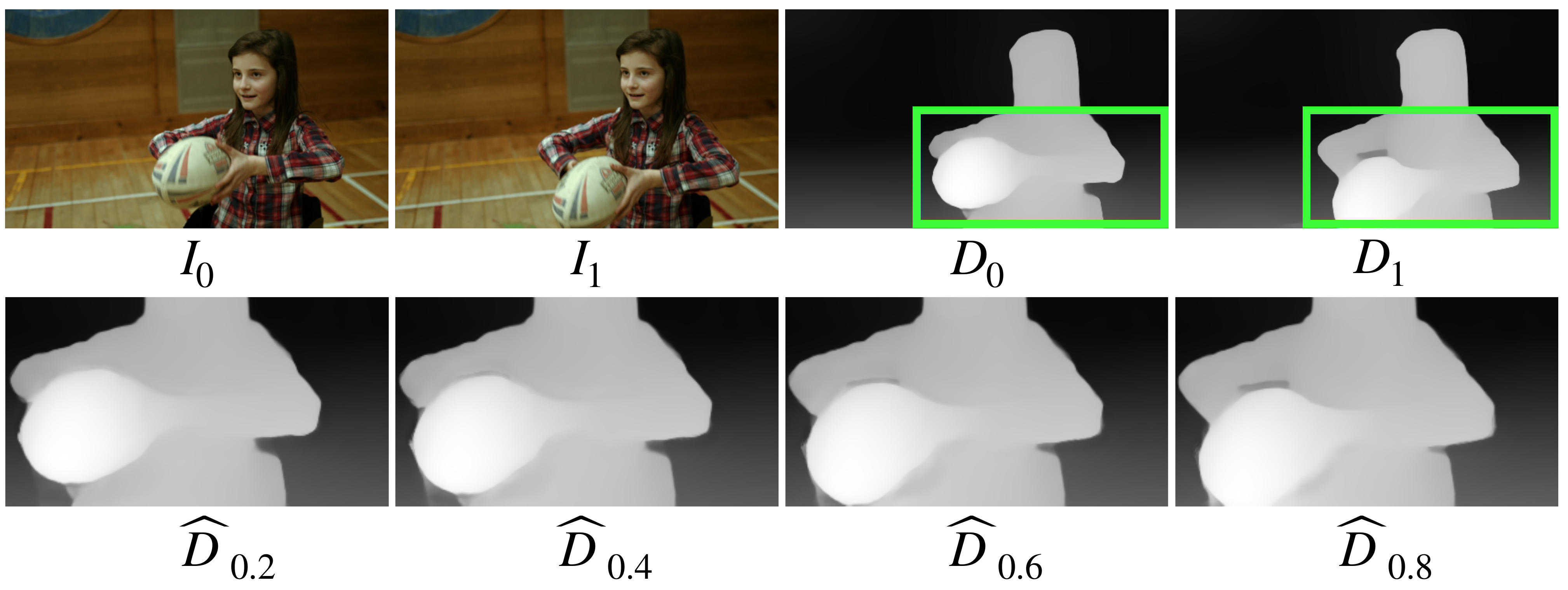}
	\vspace{-1em}
	\caption{\textbf{Interpolation for depth map using RIFE$_m$}. } 
	\label{fig:depth}
	\vspace{-0.5em}
\end{figure}

\begin{table}[t]
	\caption{\textbf{Ablation study on distillation scheme, intermediate flow estimation, model design and loss function}}
	\vspace{0.5em}
	\centering
	\resizebox{1\textwidth}{!}{\begin{tabular}{lccc}
		\hline
		\multicolumn{1}{c}{\multirow{2}{*}{Setting}} & \multicolumn{1}{c}{Vimeo90K~\cite{xue2019video}} & \multicolumn{1}{c}{MiddleBury~\cite{baker2011database}} & \multicolumn{1}{c}{Runtime} \\ 
		\multicolumn{1}{c}{}                         & \multicolumn{1}{c}{PSNR}   & \multicolumn{1}{c}{IE}     & \multicolumn{1}{c}{$640\times 480$}            \\ \hline 
		
		\multicolumn{4}{l}{\emph{Intermediate Flow Estimation}}\\
		RAFT~\cite{teed2020raft} + linear reversal~\cite{jiang2018super} & 34.68 & 2.31 & {60ms} \\
		RAFT~\cite{teed2020raft} + CNN reversal &  34.82 & 2.24 & 65ms \\
		RAFT~\cite{teed2020raft} + reversal layer~\cite{liu2020enhanced} &  35.16 & {2.04} & 101ms \\
		PWC-Net~\cite{sun2018pwc} + reversal layer~\cite{liu2020enhanced} &  {35.24} & 2.06 & 83ms \\ 
		PWC-Net~\cite{sun2018pwc} + forward warping~\cite{niklaus2020softmax} & 35.48 & {2.02} & 52ms\\
		RIFE& \textbf{35.61} & {1.96} & 16ms
		\\ \hline
		\multicolumn{4}{l}{\emph{Distillation Scheme}}\\
		RIFE w/ self-consistency & 35.37 & 2.02 & 16ms\\
		RIFE w/ RAFT-KD & {35.52} & {1.98} & 16ms\\
		RIFE~(priviledged distillation)& \textbf{35.61} & {1.96} & 16ms
		\\ \hline
		\multicolumn{4}{l}{\emph{Model Design}}\\
		RIFE w/ one IFBlock & 35.17 & 2.12 & \textbf{12ms}\\ 
		RIFE w/ two IFBlocks & {35.46} & {1.97} & 14ms\\
		RIFE + BN~\cite{ioffe2015batch} & {35.49} & {2.02} & 21ms\\
		RIFE & \textbf{35.61} & {1.96} & 16ms\\\hline
		\multicolumn{4}{l}{\emph{Loss Function}}\\
		RIFE w/ $L_1$ & {35.51} &
		\textbf{1.94} & 16ms \\
		RIFE~w/ $L_{Lap}$ & \textbf{35.61} & {1.96} & 16ms \\
		\hline 
		\normalsize
		\vspace{-1.5em}
	\end{tabular}
	}
	\label{tab:ablation2}
\end{table}

\subsection{Image Representation Interpolation}

RIFE$_m$ can interpolate other image representations using the intermediate flows and fusion map approximating from images. For instance, we interpolate the results of MiDaS~\cite{Ranftl2020} which is a popular monocular depth model, shown in Figure~\ref{fig:depth}. The synthesis formula is simply as follows:

\begin{equation}
\widehat{D}_t = M \odot \backwardwarp(D_0, F_{t\rightarrow 0}) + (1 - M) \odot \backwardwarp(D_1, F_{t\rightarrow 1}),
\end{equation}
where $D_0, D_1$ are estimated by MiDas~\cite{Ranftl2020} and $F, M$ are estimated by RIFE$_m$. RIFE may potentially be used to extend some models and provide visually
plausible effects when we ignore z-axis motion of objects.

\subsection{Ablation Studies}
\label{sec:model_ablation}


We design some ablation studies on the intermediate flow estimation, distillation scheme, model design and loss function, shown in Table~\ref{tab:ablation2}. These experiments use the same hyper-parameter setting and evaluation on Vimeo90K~\cite{xue2019video} and MiddleBury~\cite{baker2011database} benchmarks. 

\noindent\textbf{IFNet vs. Flow Reversal.} We compare IFNet with previous intermediate flow estimation methods. Specifically, we use RAFT~\cite{teed2020raft} and PWC-Net~\cite{sun2018pwc} with officially pre-trained parameters to estimate the bi-directional flows. Then we implement three flow reversal methods, including linear reversal~\cite{jiang2018super}, using a hidden convolutional layer with $128$ channels, and the flow reversal layer from EQVI~\cite{liu2020enhanced}. The optical flow models and flow reversal modules are combined together to replace the IFNet. Furthermore, we try to use 
the forward warping~\cite{niklaus2020softmax} operator to bypass flow reversal. These models are jointly fine-tuned with RefineNet. Because these models can not directly approximate the fusion map, the fusion map is subsequently approximated by RefineNet. As shown in Table~\ref{tab:ablation2}, RIFE is more efficient and gets better interpolation performance. These flow models can estimate accurate bi-directional optical flow, but the flow reversal has difficulties in dealing with the object shift problem illustrated in Figure~\ref{fig:IFFlow}. 

\noindent\textbf{Ablation on the Distillation Scheme.} We observe that removing the distillation framework makes model training sometimes divergent. Furthermore, we show the importance of distillation design in following experiments. \textbf{a1}) Remove the privileged teacher block and use the last IFBlock’s results to guide the first two IFBlocks, denoted as ``self-consistency"; \textbf{a2}) Use pre-trained RAFT~\cite{teed2020raft} to estimate the intermediate flows based on the ground truth image, denoted as ``RAFT-KD". This guidance is inspired by the pseudo-labels method~\cite{lee2013pseudo}. However, this implementation relies on the pre-trained optical flow model and extremely increases the training duration~($3\times$). We found \textbf{a1} and \textbf{a2} suffer in quality. These experiments demonstrate the importance of optical flow supervision. Some recent work~\cite{kong2022ifrnet,lu2022video} has also echoed the improvement using suitable optical flow distillation.

\noindent\textbf{Ablation on RIFE's Architecture and Loss Function.} To verify the coarse-to-fine strategy of IFNet, we removed the first IFBlock and the first two IFBlocks in two experiments, respectively. We also try some other popular techniques, such as Batch Normlization~(BN)~\cite{ioffe2015batch}. BN does stabilize the training, but degrades final performance and increases inference overhead. We provide a pair of experiments to show $L_{Lap}$~\cite{niklaus2018context,niklaus2020softmax} is quantitatively better than $\mathcal{L}_1$. 

\noindent\textbf{Limitations.} Our work may not cover some practical application requirements. Firstly, RIFE focuses on using two input frames and multi-frame input~\cite{xu2019quadratic,liu2020enhanced,kalluri2020flavr} is left to future work. One straightforward approach is to extend IFNet to use more frames as input. Secondly, most experiments are done with SSIM and PSNR as quantitative indexes. If human perception quality is preferred, RIFE can readily be changed to use the perceptually related losses~\cite{blau2018perception,niklaus2017video}. Thirdly, 
additional training data may be necessary for extending RIFE to various applications, such as interpolation for depth map and animation videos~\cite{siyao2021deep}. 
	\section{Conclusion}
We develop an efficient and flexible algorithm for VFI, namely RIFE. A separate neural module IFNet directly estimates the intermediate optical flows, supervised by a privileged distillation scheme, where the teacher model can access the ground truth intermediate frames. Experiments confirm RIFE can effectively process videos of different scenes. Furthermore, an extra input with temporal encoding enables RIFE for arbitrary-timestep frame interpolation. The lightweight nature of RIFE makes it much more accessible for downstream tasks.

\vspace{0.5em}
\noindent\textbf{Acknowledgment} 

\noindent This work is supported by National Key R\&D Program of China (2021ZD0109803) and National Natural Science Foundation of China under Grant No. 62136001, 62088102.
	
\bibliographystyle{splncs04}
\bibliography{egbib}
\section{Appendix}

\subsection{Architecture of RefineNet}
Following the previous work~\cite{niklaus2020softmax}, we design a RefineNet with an encoder-decoder architecture similar to U-Net and a context extractor. The context extractor and encoder part have similar architectures, consisting of four convolutional blocks, and each of them is composed of two $3\times 3$ convolutional layers, respectively. The decoder part in the FusionNet has four transpose convolution layers. 

\begin{figure}[h]
	\centering
	\includegraphics[width=11cm]{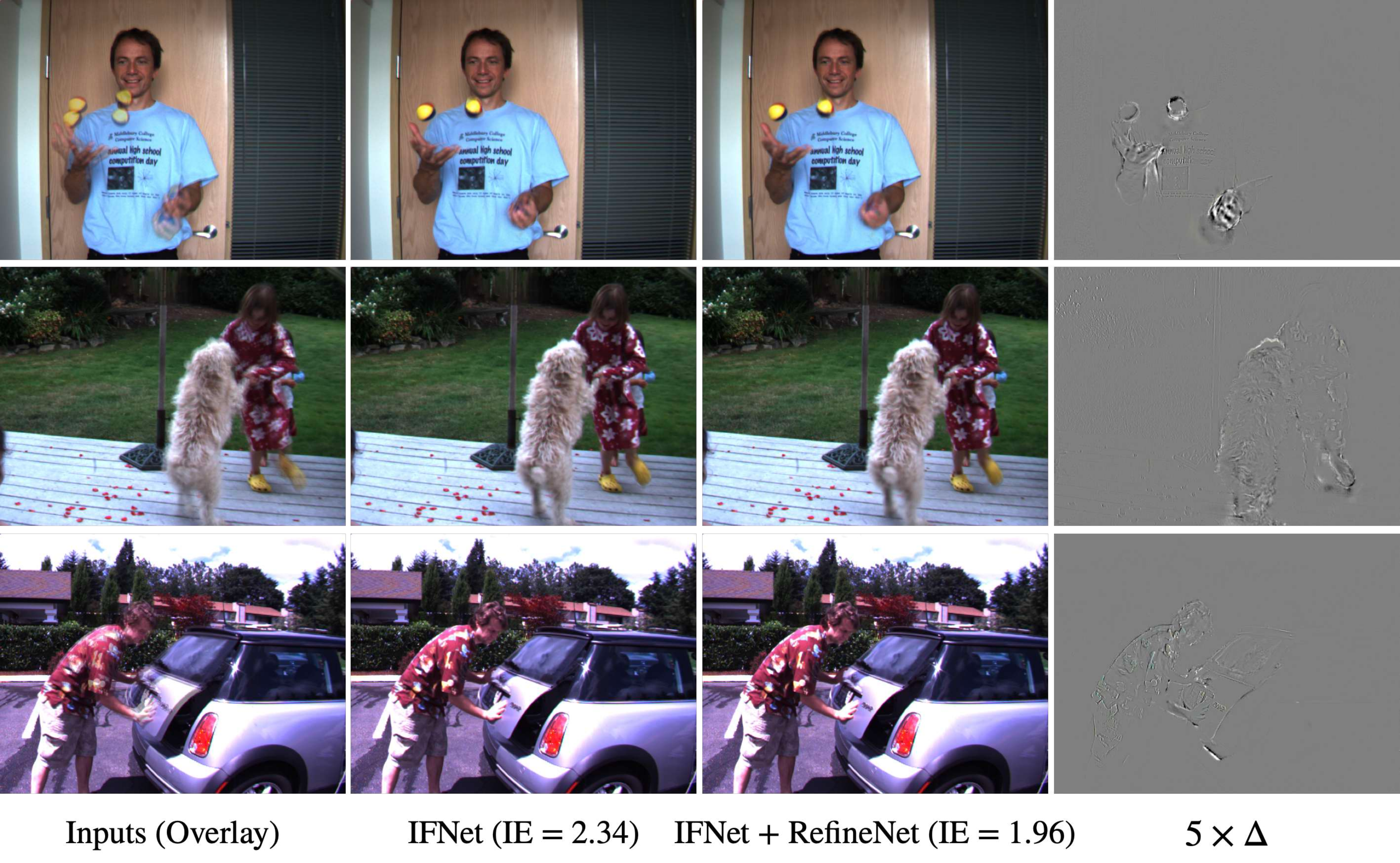}
	\caption{\textbf{Visualization of the effect of RefineNet on M.B.~\cite{baker2011database} benchmark.} }\label{fig:mb}
\end{figure}

Specifically, the context extractor first extracts the pyramid contextual features from input frames separately. We denote the pyramid contextual feature as $C_0$: $\{C_0^0, C_0^1, C_0^2, C_0^3\}$ and $C_1$: $\{C_1^0, C_1^1, C_1^2, C_1^3\}$. We then perform backward warping on these features using estimated intermediate flows to produce aligned pyramid features, $C_{t \leftarrow 0}$ and $C_{t \leftarrow 1}$. The origin frames $I_0, I_1$, warped frames $\widehat{I}_{t\leftarrow 0}, \widehat{I}_{t\leftarrow 1}$, intermediate flows $F_{t\rightarrow 0}, F_{t\rightarrow 1}$ and fusion mask $M$ are fed into the encoder. The output of $i-th$ encoder block is concatenated with the $C_{t \leftarrow 0}^{i}$ and $C_{t \leftarrow 1}^{i}$ before being fed into the next block. The decoder parts finally produce a reconstruction residual $\Delta$. And we will get a refined reconstructed image $clamp(\hat{\mathbf{I}}_t + \Delta, 0, 1)$, where $\hat{\mathbf{I}}_t$ is the reconstruct image before the RefineNet. We show some visualization results in Figure~\ref{fig:mb}. RefineNet seems to make some uncertain areas more blurred to improve quantitative results. 

\begin{figure}[h]
	\centering
	\includegraphics[width=10cm]{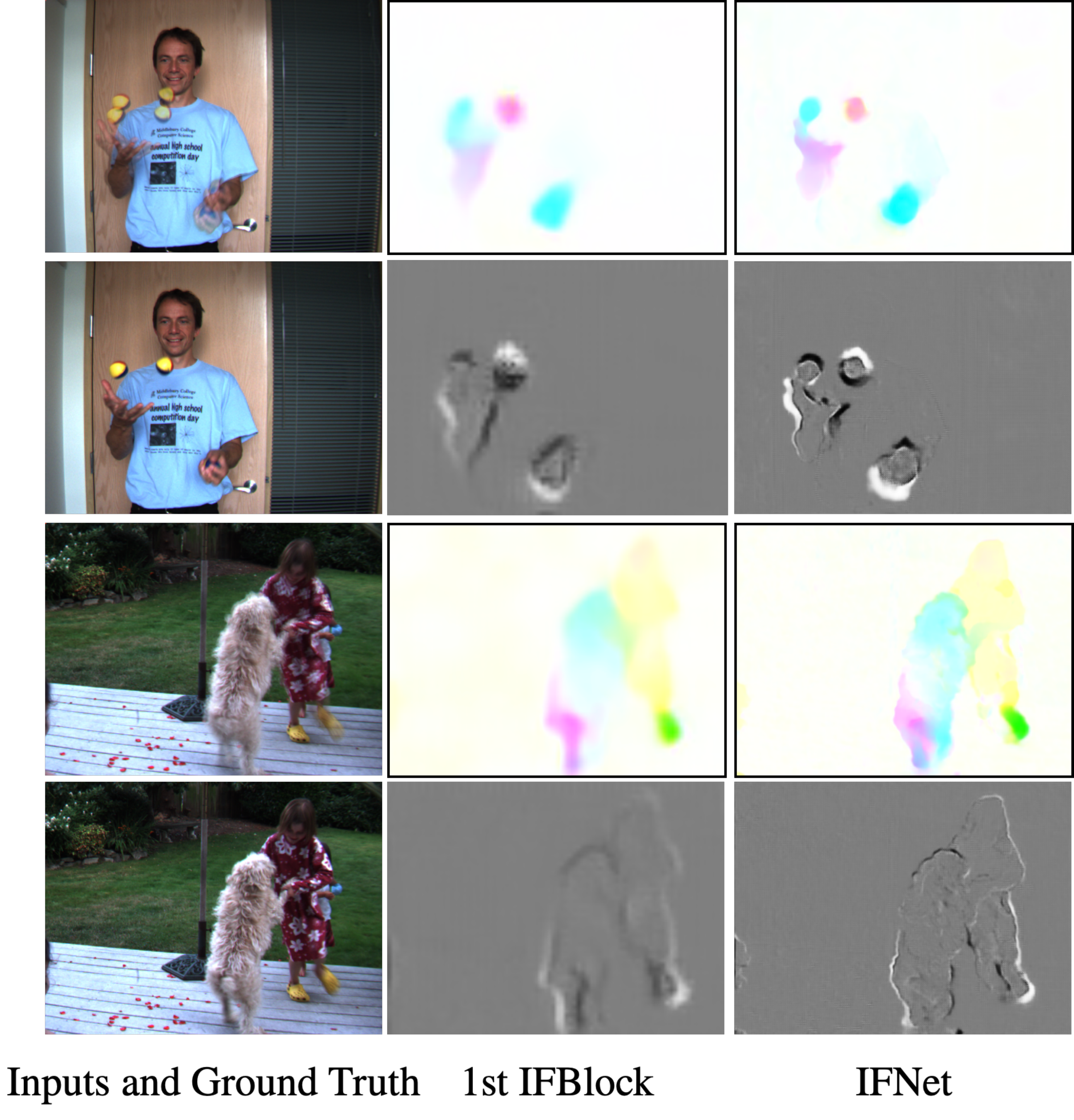}
	\caption{\textbf{Visualization of intermediate flow $F_{t\rightarrow 0}$ and blend mask $M$.} We show that stack 3 IFblocks can get finer intermediate flow and blend mask.  }\label{fig:if}
\end{figure}

\subsection{Selection of Building Operators}
We focus on introducing a simplified VFI pipeline without bells and whistles. So we choose building operators with intentional restraint. Exploring model compression is orthogonal to our approach. Our pipeline can be further sped up by manual model design, or Neural Architecture Search~(NAS) approaches (left to future work). Furthermore, plain Conv is highly supported by NPU embedded in display devices and provides convenience for customized requirements.

\subsection{Intermediate Flow Visualization}
\begin{figure}[h]
	\centering
	\begin{minipage}[t]{0.3\linewidth}
		\centering
		\includegraphics[width=1\linewidth]{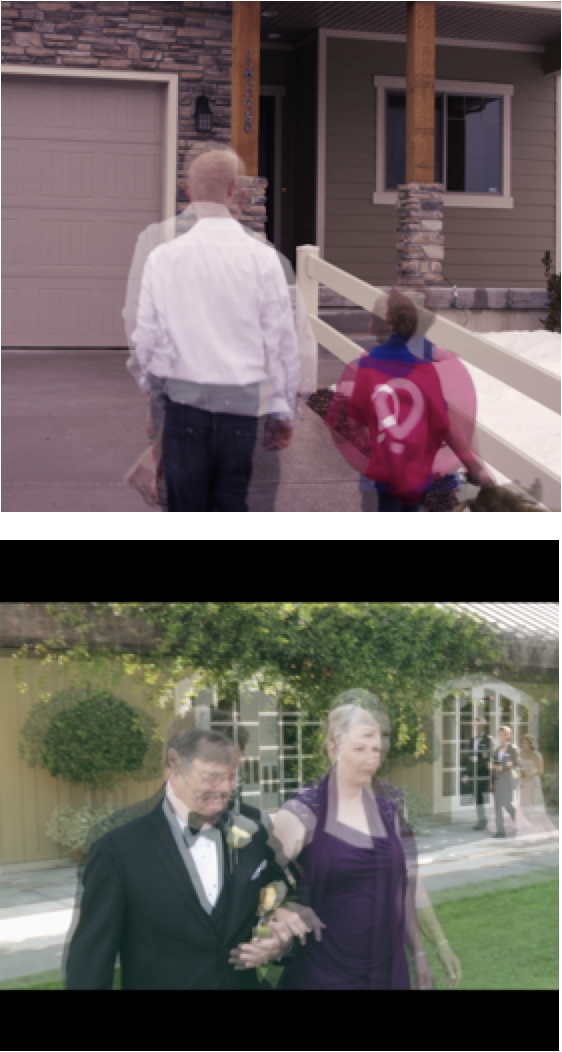}
		\centerline{Inputs~(Overlay)}
	\end{minipage}
	\begin{minipage}[t]{0.3\linewidth}
		\centering
		\includegraphics[width=1\linewidth]{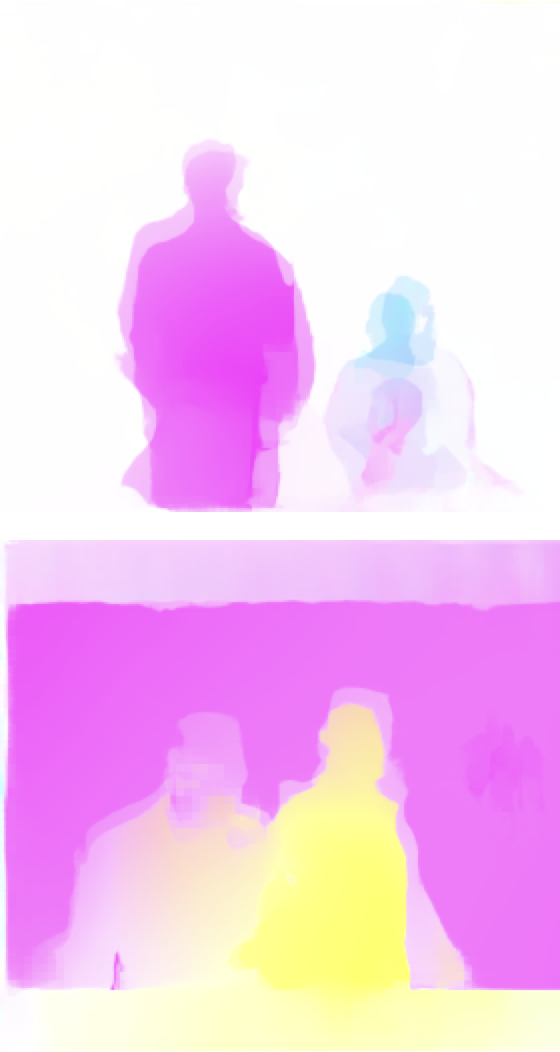}
		\centerline{Combination}
	\end{minipage}
	\begin{minipage}[t]{0.3\linewidth}
		\centering
		\includegraphics[width=1\linewidth]{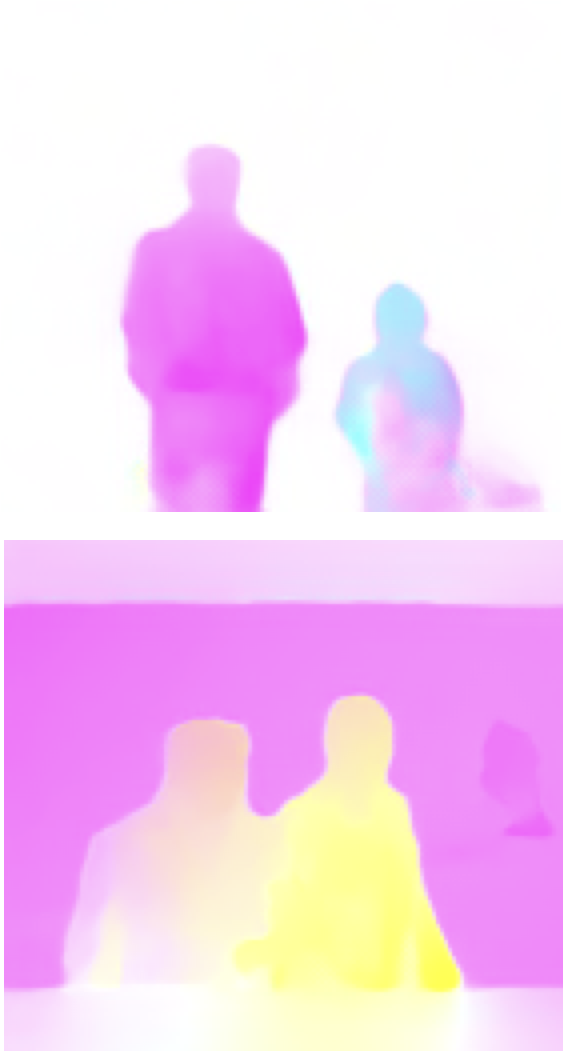}
		\centerline{IFNet}
	\end{minipage}
	
	\caption{\textbf{Visual comparison between linearly combined bi-directional flows~\cite{jiang2018super} and the result of IFNet.} }
	\label{fig:compare_flow}
	\vspace{-1em}
\end{figure}

In Figure~\ref{fig:compare_flow}, we provide visual results of our IFNet and compare them with the linearly combined bi-directional optical flows~\cite{jiang2018super}. IFNet produces clear motion boundaries.

\subsection{Model Efficiency Comparison}
Recall that we aim to explore real-time models instead of refreshing SOTA with larger models. Our models are suitable for real-time processing scenarios~(display devices, live streaming, games) and media post-processing. However, to the best of our knowledge, currently published papers do not test the speed of each state-of-the-art VFI model on same hardware, and rarely report the complexity of the model. Some previous works~\cite{bao2019depth,park2021ABME} report runtime data from the MiddleBury public leaderboard without indicating running devices. These data are reported by the submitters of various methods. A more verifiable survey comes from EDSC (Table 10)~\cite{cheng2020multiple}. We collect the models of each paper and test them on a NVIDIA TITAN X(Pascal) GPU with same hardware. The code can be found on \url{https://github.com/megvii-research/ECCV2022-RIFE/blob/main/benchmark/testtime.py}. RIFE use $16ms$ for interpolating a $640\times 480$ frame, $31ms$ for 720p frame, $68ms$ for 1080p frame. The relationship between runtime and resolution is roughly linear.

Take into account the imprecise comparison issues that TTA may introduce. We can use other techniques to get large models. We replace TTA with ``multiply the number of hidden filter's channel by a factor of $1.5$". And we train this new large model from scratch. The difference between its performance and RIFE-Large is almost negligible. $2\times$ TTA do not change the performance curve of our model. Using TTA, we can get a larger model without training.

\subsection{Other Details}
\begin{figure}[ht]
	\centering
	\includegraphics[width=7cm]{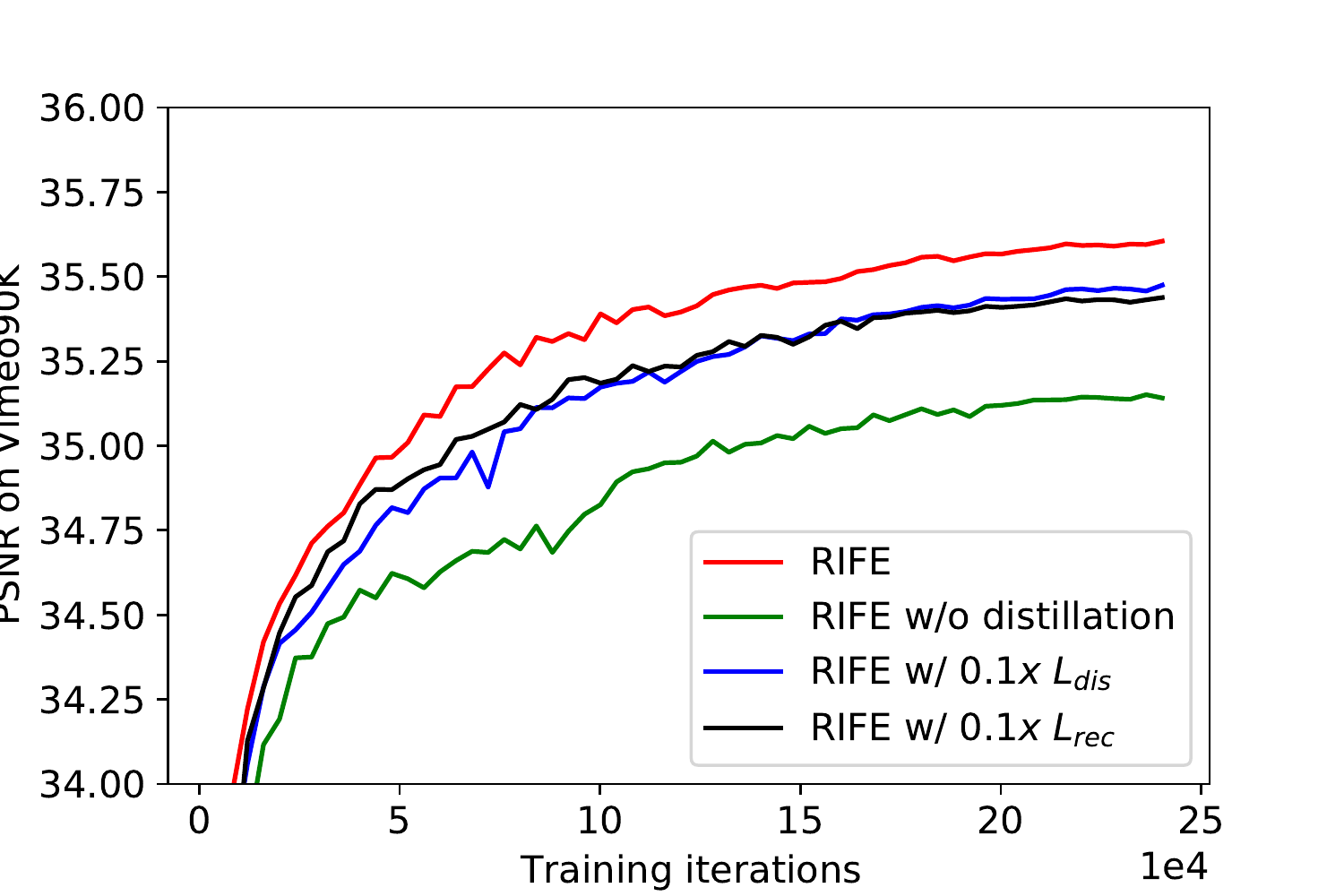}
	\caption{\textbf{PSNR on Vimeo90K benchmark during the whole training process. } The distillation scheme helps RIFE converge to better performance}\label{fig:dynamic}
\end{figure}

\noindent \textbf{Training dynamic.} We study the dynamic during the RIFE training. As shown in Figure~\ref{fig:dynamic}, the privileged distillation scheme helps RIFE converge to better performance. Furthermore, we try to adjust the weights of losses. We found that larger scale$~(10\times)$ of weights will cause the model to not converge and smaller weights$~(0.1\times)$ will slightly reduce model performance. We found that the effect of our distillation method is similar to regularization techniques, making the models easier to train.

\noindent \textbf{Supervision of mask}.  Further experiment show that adding supervision of fusion mask has no effect. More detailed distillation design may be a future research direction. When fixing $\widehat{I}_{t\leftarrow 0}$ and $\widehat{I}_{t\leftarrow 1}$, $M$ can be directly learned from the ground truth using the reconstruction loss. 
\end{document}